\documentclass{article} 
\usepackage{iclr2026_conference,times}

\usepackage{amsmath,amsfonts,bm}

\def\eqref#1{equation~\ref{#1}}

\def\1{\bm{1}}

\DeclareMathAlphabet{\mathsfit}{\encodingdefault}{\sfdefault}{m}{sl}
\SetMathAlphabet{\mathsfit}{bold}{\encodingdefault}{\sfdefault}{bx}{n}

\newcommand{\AlgOursAbrv}{SEED-SET}
\newcommand{\AlgOurs}{\textcolor{MaterialBlue900}{\small\textsf{\AlgOursAbrv{}}}}

\newcommand{\ouralgoname}[1]{\textcolor{MaterialBlue900}{\small\textsf{#1}}}

\newcommand{\baselinename}[1]{\textcolor{MaterialPurple900}{\textsf{#1}}}
\newcommand{\taskname}[1]{\textsf{#1}}

\newcommand{\fbus}[0]{\textsf{5-Bus}}
\newcommand{\tbus}[0]{\textsf{30-Bus}}

\usepackage{hyperref}
\usepackage{url}

\usepackage[utf8]{inputenc} 
\usepackage[T1]{fontenc}    
\usepackage{booktabs}       
\usepackage{amsfonts}       
\usepackage{nicefrac}       
\usepackage{microtype}      
\usepackage[dvipsnames]{xcolor}         
\usepackage{graphicx}
\usepackage{amsmath}
\usepackage{amssymb}
\usepackage{mathtools}
\usepackage{amsthm}
\usepackage{comment}
\usepackage{caption}
\usepackage{subcaption}
\usepackage{doi}

\usepackage{enumitem}
\usepackage{hyperref} 
\usepackage[nameinlink,capitalise,noabbrev]{cleveref} 
\usepackage{xcolor-material}
\usepackage{wrapfig}
\usepackage{tcolorbox}
\tcbuselibrary{skins, breakable}

\newenvironment{promptcard}[1]{
\begin{tcolorbox}[
  breakable,
  enhanced,
  sharp corners,
  colback=gray!5,
  colframe=black!40,
  title={#1},
  fonttitle=\bfseries,
  left=3mm, right=3mm, top=2mm, bottom=2mm
]}
{\end{tcolorbox}}

\definecolor{MajorEditColor}{HTML}{CC2020}

\theoremstyle{plain}
\newtheorem{problem}{Problem}[section]

\title{SEED-SET: Scalable Evolving Experimental Design for System-level Ethical Testing}

\author{
Anjali Parashar$^{1}$\thanks{Corresponding author} 
\quad Yingke Li$^{1}$
\quad Eric Yang Yu$^{1}$
\quad Fei Chen$^{1}$ \\
\textbf{James Neidhoefer$^{1}$}
\quad \textbf{Devesh Upadhyay$^{2}$}
\quad \textbf{Chuchu Fan$^{1}$}
\\
$^{1}$Laboratory for Information \& Decision Systems (LIDS), MIT 
\quad \quad $^{2}$Saab Inc. \\
\texttt{\{anjalip,yingkeli,eyyu,feic,jimneid,chuchu\}@mit.edu}\\
\texttt{devesh.upadhyay@saabinc.com} 
}
\iclrfinalcopy
\begin{document}

\maketitle

\begin{abstract}

    As autonomous systems such as drones, become increasingly deployed in high-stakes, human-centric domains, it is critical to evaluate the ethical alignment since failure to do so imposes imminent danger to human lives, and long term bias in decision-making. Automated ethical benchmarking of these systems is understudied due to the lack of ubiquitous, well-defined metrics for evaluation, and stakeholder-specific subjectivity, which cannot be modeled analytically.  To address these challenges, we propose SEED-SET, a Bayesian experimental design framework that incorporates domain-specific objective evaluations, and subjective value judgments from stakeholders. SEED-SET models both evaluation types separately with hierarchical Gaussian Processes, and uses a novel acquisition strategy to propose interesting test candidates based on learnt qualitative preferences and objectives that align with the stakeholder preferences.
    We validate our approach for ethical benchmarking of autonomous agents on two applications and find our method to perform the best. Our method provides an interpretable and efficient trade-off between exploration and exploitation, by generating up to $2\times$ optimal test candidates compared to baselines, with $1.25\times$ improvement in coverage of high dimensional search spaces \footnote{Project website: \url{https://anjaliparashar.github.io/seed-site/}}.  

\end{abstract}

\section{Introduction}\label{sec:intro}
Artificial intelligence (AI)-enabled autonomous systems have seen increased deployment across a wide range of high-stakes domains, including automated energy distribution, disaster management ~\citep{battistuzzi2021ethical}.
Although such applications can bring significant social benefits~\citep{maslej2025artificial,zeng2024air,weidinger2021ethical,birhane2024dark}, they raise equally urgent ethical concerns~\citep{sovacool2016energy,bhattacharya2024trust,amodei2016concrete,jobin2019global,wang2025measuring,grabb2024risks,palka2023ai} across stakeholder groups.
For example, in the power grid resource allocation problem, energy distribution policies often prioritize higher-income areas during peak demand periods, leaving marginalized populations more vulnerable to outages~\citep{fahmin2024eco,chitikena2023robotics,cong2022energyequitygap}. 
Such examples highlight three core challenges of ethical evaluation in real-world autonomous systems:
\begin{itemize}[leftmargin=1em,itemsep=0pt]
    \item \textit{Measuring ethical behavior is difficult.} Standard ethical evaluation metrics such as fairness and social acceptability often lack ground-truth labels~\citep{mittelstadt2016ethics,salaudeen2025measurement,reuel2024open,wallach2025position}. 
    \item \textit{Value alignment is user-dependent, and evolving.} Evaluation standards must quickly adapt to the growing capabilities of autonomous systems~\citep{keswani2024pros,tarsney2024moral}. Static evaluation standards such as test suites and benchmarks require persistent revisions. Additionally, a wide range of ethical benchmarking problems are user-specific and user evaluation can be noisy.
    \item \textit{Ethical evaluation of real-world platforms is expensive.} Due to resource constraints such as budget, real-world systems require sample-efficient evaluation. Disproportionate access to large-scale human feedback across domains also imposes sample restrictions on stakeholders. 
\end{itemize}
\begin{figure}
    \centering
    \includegraphics[width=0.9\linewidth]{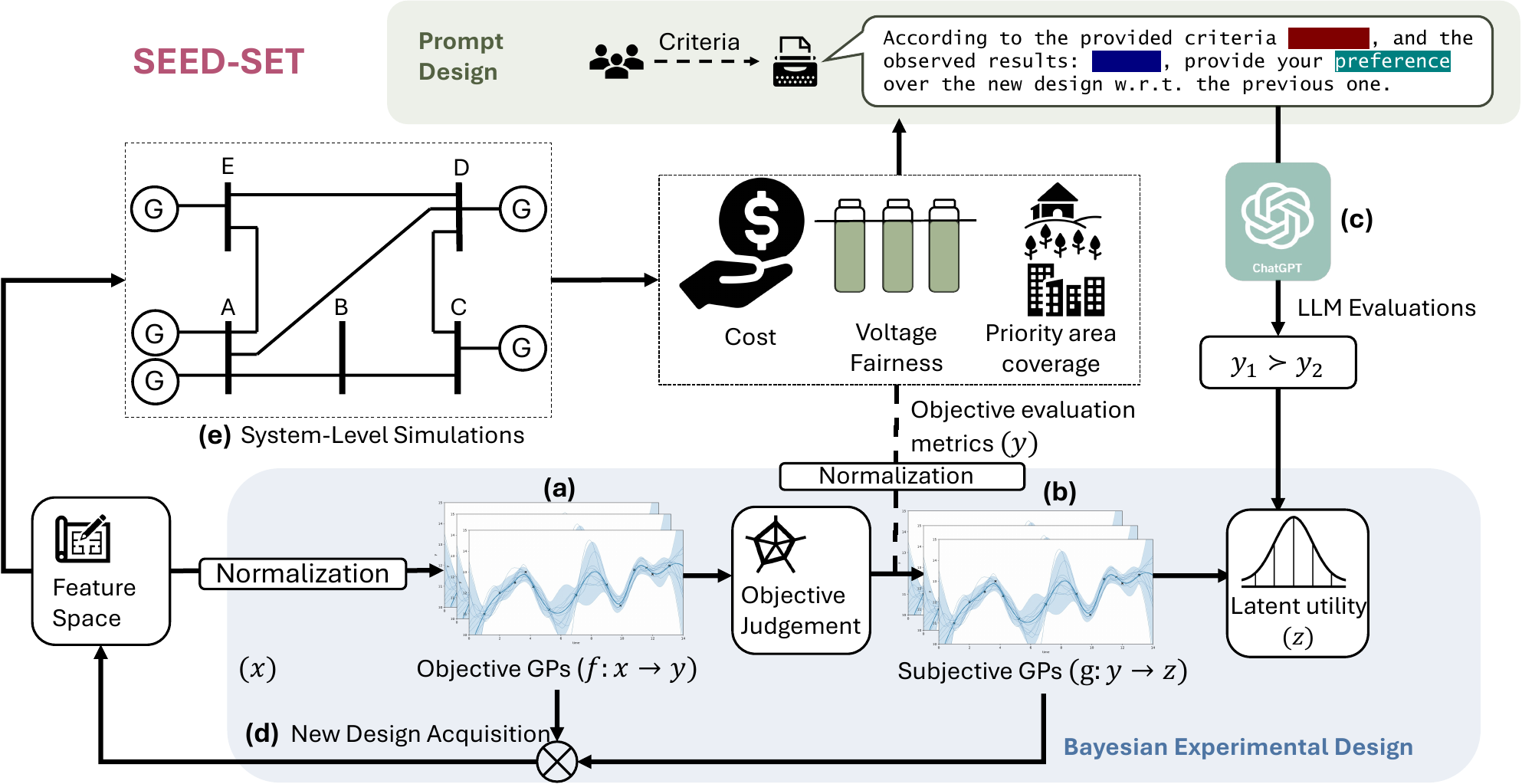}
    \caption{\textbf{SEED-SET Overview.} Our framework integrates quantitative metrics learned using Objective GP (a) with user preferences learned via a Subjective GP through pairwise elicitation (b). An LLM performs pairwise comparisons of scenario outcomes (c) to inform the acquisition process (d), which generates a pair of scenarios for evaluation, aligned with user-defined ethical criteria. These scenarios are then used for system-level simulations (e) in a sequential manner.}
    \label{fig:framework}
\end{figure}
To address the first challenge, guidelines and standards for ethical behavior in AI systems have been proposed~\citep{AI-RMF,IEEE-P7001,ISO/PAS-8800,chance2023assessing}. For example, NIST’s AI Risk Management Framework (AI RMF 1.0, 2023)  that suggests high-level guidelines (e.g., Govern, Map, Measure, Manage) to promote `trustworthy AI'. Although these guidelines serve as useful heuristics for enforcing measurable ethical behavior, they are not sufficiently concrete for direct system-level testing. 
Some recent efforts in this direction have led to automated evaluation tools, largely focusing on rule-based ethical benchmarking based solely on established guidelines~\citep{reuel2022using,dennis2016formal}, or preference-based methods using human feedback~\citep{keswani2024pros,liu2024aligning}, and often argue in favor of one over the other. Instead, we argue that in its most general form, ethical evaluation must incorporate objective feedback from existing guidelines, as well as stakeholder concerns.

Additionally, existing works assume abundant access to cheap simulations or expert annotations, leading to sample-extensive approaches based on reinforcement learning (RL), reinforcement learning from human feedback (RLHF), or adaptive stress testing~\citep{reuel2022using,dennis2016formal,gao2024ethical}. Such assumptions restrict their applicability to real-world systems, underscoring the need for methods that unify both forms of evaluation under realistic data and resource constraints.

 To address this, we propose \textbf{Scalable Evolving Experimental Design for System-level Ethical Testing (\AlgOursAbrv{})}, an evaluation methodology that benchmarks autonomous systems against both objective measurable metrics (e.g. fire damage to buildings) and subjective ethical metrics (e.g. rescue priority to different vulnerable groups) while maintaining low sampling requirements. \Cref{fig:framework} provides an overview of our method.

To our knowledge, this is the first framework of its kind to explicitly consider both objective and subjective ethical evaluation criteria. A key nuance in this design is the interplay of objective metrics and stakeholder preferences. Stakeholder preferences are affected by and dictate regions of importance in the objective landscape, and strategy to explore these regions in the objective metrics must adapt to stakeholders. This dual dependency is highly non-trivial, and not explicitly acknowledged in prior works. We incorporate this dependency in the design of our novel data acquisition strategy, that incorporates feedback from both models for proposing challenging test cases (~\Cref{sec:seed}). We evaluate SEED-SET on three tasks for ethical evaluation: Power system resource allocation, Fire rescue by aerial autonomous agents, and Optimal route design in urban traffic (~\Cref{sec:experiment}). Our method successfully scales to high dimensional scenarios, by proposing up to $2\times$ more optimal test cases compared to baselines.

Our contributions can be summarized as follows:

\begin{itemize}[leftmargin=1em,itemsep=0pt]
    \item We introduce a unified, domain-agnostic problem formulation for system‑level ethical testing, modeling it as an adaptive, sample‑constrained inference task over both objective metrics and subjective values.
    \item We formalize a hierarchical Variational Gaussian Process (VGP) model that maps design parameters to measurable ethical criteria and learns their utility according to subjective factors.
    \item  We derive a novel joint acquisition criterion for hierarchical models that balances exploration of uncertain ethical factors with exploitation of learned ethical preferences in our hierarchical VGP.
\end{itemize}

\section{Related Work}\label{sec:relatedwork}
We discuss key related works here, with a detailed discussion in ~\Cref{sec:lit}.


\textbf{Governance approaches for responsible AI.} A wide range of governance frameworks, guidelines and standards have been proposed to guide the ethical development and deployment of AI systems \citep{AI-RMF, oecd2019principles, ieee2019aligned,IEEE-P7001,ISO/PAS-8800}). These works articulate high-level values but are vague about specific mechanisms for practical enforcement \citep{hagendorff2020ethics}. For domain-specific implementation of these guidelines, automated ethical evaluation tools have been proposed in the literature.

\textbf{Automated tools.} Prior technical approaches include reinforcement learning and orchestration for instilling ethical values~\citep{noothigattu2019teaching}, large-scale studies of moral judgment in LLMs~\citep{zaim2025large}, and active learning for preference elicitation~\citep{keswani2024pros}. With the exception of active learning, these techniques impose large-scale data and simulation budget requirements and lack interpretablity and modularity provided by our framework. 


\section{Problem Statement}\label{sec:setup}

We formulate system-level ethical testing as follows:

\begin{problem}
    Given a black-box autonomous system $\mathcal{S}_{\pi}$, parameterized by policy $\pi \in \Pi$ (such as power resource allocation, drone navigation in environment), evaluate its ethical alignment by querying it in scenarios $x\in \mathcal{X}$ (environment properties such as location of assets), collecting objective evaluations $y \in \mathcal{Y}$ (such as cost, resilience), and estimating an unknown ethical compliance function $f:\Pi \times \mathcal{X} \rightarrow \mathbb{R}$ that captures both objective outcomes and subjective value judgments.
\end{problem}

We list some design choices to meet the key requirements of our problem formulation: 

\textbf{Multi-faceted ethical criteria.} The overall subjective evaluation depends on some task-specific, and some task-agnostic parameters. For example, a stakeholder in power resource allocation will prefer low cost and high grid reliability, regardless of the system specifics, such as grid size. Thus, we decompose ethical compliance $f(\pi,x)$ into two parts: a set of objective metrics $f_{\text{obj}}:\mathcal{X}\to\mathcal{Y}$ (e.g., cost, reslience) which can be modeled analytically using prior knowledge (domain experts, guidelines) and user specific subjective evaluation $f_{\text{subj}}:\mathcal{Y} \to \mathbb{R}$ (e.g., perceived fairness using cost, resilience as metrics), with limited access to ground-truth evaluations. For a given $y$, $f_{\text{subj}}(y)$ represents the degree of ethical alignment with the subjective evaluation criteria. 


\textbf{Sample-constrained learning.} Evaluation is costly: querying $\mathcal{S}_{\pi}$ in scenario $x$ incurs a cost $c(\pi,x)$. We approach this using a sequential design paradigm. Given a total testing budget $B$, the goal of Bayesian Experimental Design is to sequentially select query points to best learn $f$ within budget, to maximize the amount of information obtained about the model parameters of interest. This promotes sample efficiency by utilizing information from collected data $\mathcal{D}:=\{(x_{1},y_{1}), \dots, (x_{n},y_{n})\}$. 


\textbf{Scalability and uncertainty modeling.} The space $\Pi \times \mathcal{X}$ may be high-dimensional, and complex, and evaluations may inherit uncertainty from experts and stakeholders. We model this uncertainty cumulatively in the objective evaluation as $y \sim \mathcal{N}(f(x),\sigma^{2}(x))$, by assuming the noise follows a zero-mean normal distribution with a standard deviation $\sigma$. Ethical testing thus requires  explicit uncertainty modeling (e.g., via Bayesian inference) and scalable function approximation (e.g., variational models), to guide testing toward the most informative scenarios.

\textbf{Assumptions.}
We make the following assumptions about the system under test $\mathcal{S}_{\pi}$, and the user:

\begin{enumerate}[label=\textbf{A\arabic*},leftmargin=1em,itemsep=0pt]
    \item The policy $\pi$ is fixed during testing, and the scenario space $\mathcal{X}$ is known and fixed a priori.
    \item The user provides their true subjective ethical preferences (i.e., users do not misreport), as the ethical evaluation depends on the user-defined notion of “good” or “preferred” behavior. Additionally, we assume that given a set of objectives, the user's latent ethical model is stationary, corresponding to an unknown but fixed subjective criterion.
    \item We additionally assume access to objective evaluations (e.g., damage caused by fire), which are required for subjective evaluation. The complete set of objective metrics is assumed to be fully known a priori.
\end{enumerate}


Assumption A1 pertains to system requirements, and is a widely used, fundamental assumption in BO/BED literature~\citep{rainforth2024modern,Chaloner1995BayesianReview}. This is needed to ensure a fixed feature space for learning surrogate models for objective and subjective criteria. 

Assumption A2 is rooted in pairwise elicitation literature that assumes a stationary latent utility function associated to preferential evaluation~\citep{Chu2005PreferenceProcesses}. Note that our framework still accommodates stochasticity of evaluation using a probabilistic framework, assuming a stationary latent utility model. 

Assumption A3 lists assumptions on objectives, and is also made in prior works investigating composite optimization and preferential evaluation using human feedback~\citep{Christiano2017DeepRL,lin2022preference,astudillo2019bayesian}, which consider human feedback over explicitly available `observables' generated from a system. This also mirrors real-world preference elicitation settings, where user judgments are anchored to clearly defined objective attributes~\citep{shao2023eliciting}.

Following the hierarchical distinction of $f$, we meet the multifaceted ethical evaluation requirement using the design of a hierarchical surrogate model, learnt using data queried by our proposed acquisition strategy, in a Variational Bayesian Experimental Design loop. Furthermore, we mitigate the scalability constraints with human evaluation using LLM as proxy evaluators, for a fixed ethical criterion. We now discuss the specifics of our methodology. 



\section{SEED-SET}\label{sec:seed}
Our approach consists of three main modeling components, a hierarchical VGP for surrogate modeling, a data acquisition strategy to generate test cases, combined with an LLM as a proxy evaluator for pairwise preferential evaluation. ~\Cref{fig:framework} provides an overview of our approach.  

\subsection{A Variational Bayesian Framework for Scalable Ethical Modeling}\label{subsec:problem}

Our three main components interact with each other with inherent stochasticity from system observations and uncertainty from limited user evaluations. To account for these considerations in a sample-efficient evaluation setting, we adopt a variational Bayesian framework.  Specifically, we learn a surrogate model for $f$ using $f(x) \sim p(f(x)|\mathcal{D})$ by applying a joint distribution over its behavior at each sample $x \in \mathcal{X}$. The prior distribution of the objective $p(f(x))$ is combined with the likelihood function $p(\mathcal{D}|f(x))$ to compute the posterior distribution $p(f(x)|\mathcal{D})\propto p(\mathcal{D}|f(x))p(f(x))$, which represents the updated beliefs about $f(x)$. We approximate the posterior $p(f(x)|\mathcal{D})$ using a variational distribution parameterized by $\phi$, as $q_{\phi}(f(x))$, for sample-efficient posterior estimation.  

In this work, we use GPs to estimate the posterior distribution, due to their analytical compatibility with evaluation under limited data. In GP models, the distribution is a joint normal distribution $p(f(x)|\mathcal{D})=\mathcal{N}(\mu(x),k(x, x'))$ completely specified by its mean $\mu(x)$ and kernel function $k(x, x')$, where $\mu(x)$ represents the prediction and $k(x, x')$ the associated uncertainty. The computational complexity of GP models scales with $\mathcal{O}(n^{3})$ as the number of observations $n$ increases. To ensure scalability with the number of observations, we generalize our variational posterior models $q_{\phi}$ to Variational GPs (VGPs) ~\citep{tran2015variational}, that reduce the computational burden of inference through sparse approximation of the posterior distribution. A detailed discussion on the computational efficiency of VGPs is provided in ~\Cref{subsec:VGP}.

In this way, the complexity of inference is reduced from $\mathcal{O}(n^{3})$ to $\mathcal{O}(nm^{2})$, which significantly improves the efficiency if $m \ll n$.

\textbf{Hierarchical VGP (HVGP) for Modular Modelling.}
Ethical evaluation in autonomous systems is inherently hierarchical: system designs give rise to observable behaviors measured by $f_{\text{obj}}$, which in turn elicit subjective ethical evaluations $f_{\text{subj}}$ from user. To capture this structure in a scalable and interpretable way, we decompose the ethical evaluation task into two distinct modeling stages, each represented by a VGP:
\begin{itemize}[leftmargin=1em,itemsep=0pt]
    \item \textit{Objective GP}, which models the mapping  $f_{\text{obj}}$ using a surrogate $g:x\rightarrow y$, where $y \in \mathbb{R}^{d}$ are objective metrics, intermediate quantities that reflect system behaviors relevant to ethical concerns (e.g., cost of decision-making, resilience, equity in resource distribution across stakeholders, etc.).
    \item \textit{Subjective GP}, which models the mapping $f_{\text{subj}}$ using a surrogate $h:y\rightarrow z$, where $z \in \mathbb{R}$ denotes a latent utility score representing stakeholder judgments (e.g., perceived fairness or acceptability), obtained from qualitative evaluations. 
\end{itemize}



Subjective ethical evaluation lacks ground-truth values, i.e., we do not have direct access to label $z$. This makes supervised training infeasible. We therefore adopt a common practice for grounding qualitative information involves preference elicitation using pairwise evaluation from an oracle $\mathcal{T}:(y,y')\to\{1,2\}$~\citep{huang2025acceleratedpreferenceelicitationllmbased}, which are objectives from scenarios $(x,x')$. Here, the oracle takes a pair of evaluations $y,y' \in \mathcal{Y}$, and returns a binary label $``1"$ or  $``2"$, indicating its preferred design ($``1"$ if $y \succ y'$ and vice-versa). 

This hierarchical structure offers two critical advantages: 1) \textit{Interpretability}: Ethical preferences are grounded in observable system outcomes $y$, not the latent design parameters $x$. Modeling $h(y)$ instead of $h(x)$ aligns with how stakeholders assess ethical outcomes in terms of behaviors they can perceive and evaluate.   
2) \textit{Data efficiency}:  
By incorporating the subjective criteria's dependency on a combination of task-specific and task-agnostic objectives, we promote accurate modeling choices. For accurate evaluation in limited evaluations, sample efficiency and quality of evaluation is highly sensitive to modeling choices~\citep{keswani2024pros}. 


Efficient data querying is critical under limited budgets. Naive random sampling wastes evaluations and often misses key test cases. Moreover, objectives both shape and are shaped by subjective evaluations, making separate model training ineffective. Instead, one must target regions of objective space aligned with subjective criteria. We address this through adaptive data acquisition within a Bayesian Experimental Design (BED) framework.

\subsection{A Bayesian Experimental Design Loop for Adaptive and Efficient Testing}\label{subsec:acf}


Given a history of experiments $\mathcal{D}:=\{(x_{1},y_{1}), \dots, (x_{n},y_{n})\}$, BED seeks to maximize the \textit{Expected Information Gain} (EIG) that a potential experimental outcome can provide about  the model parameter of interest, denoted as $\theta$. This is measured as the expected reduction in entropy $H(\cdot)$ of the posterior distribution of $\theta$: 
\begin{equation}
    \text{EIG}(x)=H[p(\theta|\mathcal{D})] - \mathbb{E}_{p(y|x, \mathcal{D})}[H[p(\theta|\mathcal{D}\cup(x,y))]]=I(\theta;(x,y)|\mathcal{D}),
\end{equation}
where, $I$ denotes the mutual information between $\theta$ and $(x,y)$.

We generalize this paradigm into our HVGP models and propose a nested approach that unifies the exploration and exploitation of both the objective factors and subjective preferences simultaneously. We consider two variational distributions $q_{\phi}(\cdot)$ and $q_{\psi}(\cdot)$, of the Objective and Subjective GP, parameterized by $\phi,\psi$ respectively. We propose $V:\mathcal{X}\to \mathbb{R}$:
\begin{equation}\label{eq:abalation_aqui}
    \begin{split}
        V(x) = I(g_{x};y|\mathcal{D}) +\mathbb{E}_{q_{\phi}(y|x)}[I(h_{y};z|\mathcal{D})+\mathbb{E}_{q_{\psi}(h_{y})} [h_y]],
    \end{split}
\end{equation}
such that two jointly evaluated candidates $x=[x_{1},x_{2}]$ can be obtained by maximizing V. Here $g_x, h_y$ denote $g(x)$ and $h(y)$ respectively. Maximizing the first two terms maximizes mutual information in scenario and objective spaces, while the third enforces preferential alignment with proposed criteria. 

\Cref{eq:abalation_aqui} reflects the inherent hierarchical structure necessary for accurately modeling ethical preferences. The first term captures the expected information gain about the \textbf{objective layer}, ensuring that we reduce uncertainty in the objectives. The second term quantifies information gain in the \textbf{subjective layer}, directly improving our estimate of the latent utility function $h(y)$. The final term encourages sampling in regions where the current model predicts higher ethical utility, enabling the method to balance exploration with targeted exploitation. Balanced exploration–exploitation requires all three. In ~\Cref{sec:experiment}, we study acquisition ablations, and in ~\Cref{subsec:power}, stakeholder ablations on the learned objective space. In \Cref{sec:power-ablation} we demonstrate the advantage of joint acquisition in sample efficiency and discovery of scenario with high alignment.

\subsection{An LLM-based Proxy for Subjective Ethical Evaluation}\label{subsec:LLM}

The pairwise elicitation oracle is modeled using humans for evaluation. However, using human experts can lead to constraints on the number of pairwise evaluations that can be performed. Additionally, getting true experts who are not biased to provide the subjective evaluation can be challenging, especially for understudied domains, and is not cost-effective. To reduce the dependency on user evaluation, we leverage LLMs as proxies to make evaluations on behalf of stakeholders according to certain stakeholder-specified criteria encoded through prompt design~\citep{huang2025acceleratedpreferenceelicitationllmbased}. Our proposed prompt design accommodates the hierarchical structure proposed so far.

\textbf{Prompt Design:} The prompt has three main parts: 1) \textit{Task description}: Specifies task-relevant contextual details, 2) \textit{Objective metrics} $(y_1,y_2)$: For two scenarios $(x_1,x_2)$, corresponding objective metrics $(y_1,y_2)$ for chosen objectives are provided for comparison, 3) \textit{Subjective criteria}: An NLP description of preference over the objective landscape, that encodes criteria for selecting the preferred candidate. These details along with response instruction are used to extract a binary preference $``1"$ if $y_1 \succ y_2$ from the LLM. We conduct ablation studies on model and temperature in \Cref{sec:llm-ablation} that validates the reliability of LLMs as proxy evaluators.

\section{Experiments}\label{sec:experiment}


\begin{figure}
    \centering
    \includegraphics[width=0.8\linewidth]{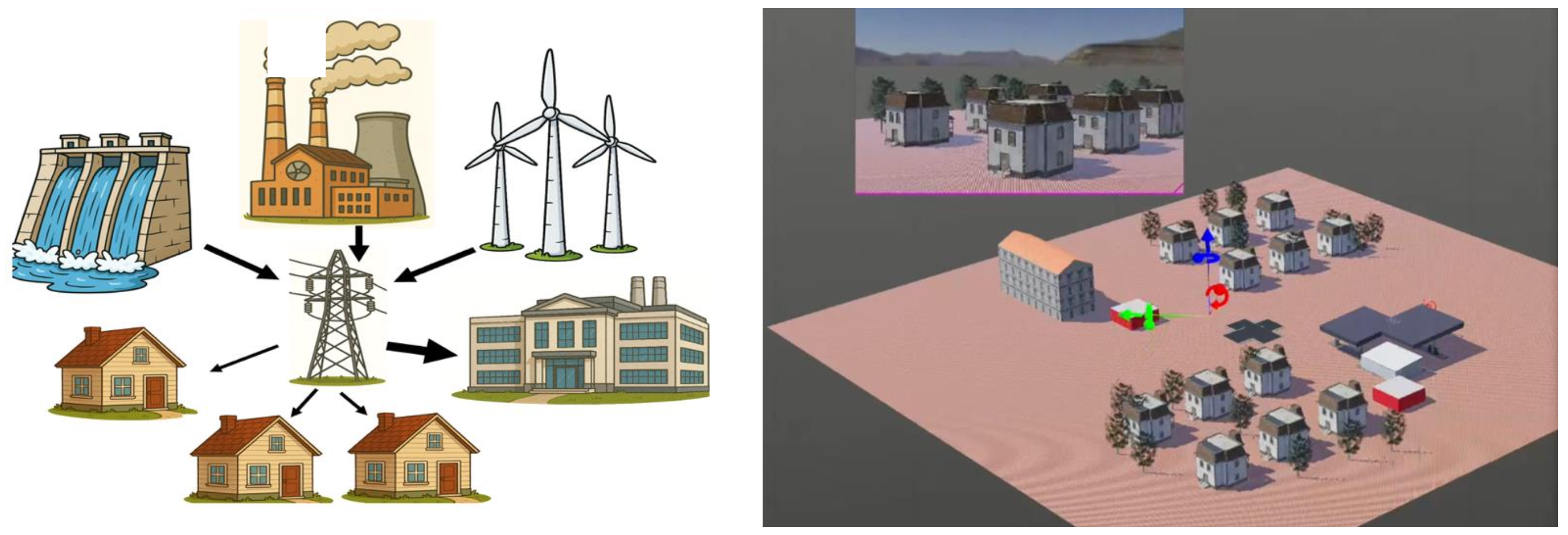}
    \caption{Environments for the two case studies considered in this work. \textbf{(Left)} Power Grid Allocation - IEEE \taskname{5-Bus} and \taskname{30-Bus} (~\Cref{subsec:power}). \textbf{(Right)} \taskname{Fire Rescue} (~\Cref{subsec:fire}). Additional case study for Optimal Routing in \Cref{appendix:optimal_routing}.}
    \label{fig:environs}
\end{figure}


Our central hypothesis is that \AlgOursAbrv{} enables scalable, accurate, and data-efficient ethical evaluation of autonomous systems.
Using previously discussed ethical evaluation concerns~\citep{battistuzzi2021ethical,luo2024ethical,NARUC_ADER_Fundamentals_2024} to guide the design of our observables, we test our hypothesis on several case studies across three main applications.
In addition, we conduct several ablations to better understand our methodology.
In all plots, solid lines are the mean $\mu$ and shaded areas are one standard deviation ($\mu \pm \sigma$).
We run for five seeds per experiment, using GPT-4o for all LLM queries. 
Additional details are provided in ~\Cref{subsec:baselines}.



\paragraph{Benchmarks.} We propose three case studies, exploring real world applications using Power Grid Resource allocation, Fire Rescue (as illustrated in ~\Cref{fig:environs}), and Optimal Routing case study (\Cref{appendix:optimal_routing}).  We also explore multi stakeholder evaluation in \Cref{sec:multi-stakeholder} and applicability of our method on real human data in \Cref{sec:travel-mode}.
To the best of our knowledge, there are no standard simulation platforms/benchmarks to test domain-agnostic ethical benchmarking techniques for low-budget experimental validation. 
 We discuss more about each case study in the coming sections.

 

\paragraph{Baselines.} We test our \AlgOurs{} framework against the following relevant baselines. 
First, the \baselinename{Random sampling} baseline samples uniformly in the design parameter space.
\baselinename{Single GP}~\citep{keswani2024pros} is a pairwise preferential GP that directly maps design parameters consisting of a pair of scenarios $(x, x^\prime)$ to ethical evaluations $z$. 
\baselinename{VS-AL-1} and \baselinename{VS-AL-2} are Version Space Active Learning baselines, referenced in \citep{keswani2024pros} that use a Support Vector Machine (SVM) to learn a preferential decision boundary for pairs of scenarios $(x,x^\prime)$, with a linear kernel and RBF kernel respectively. 
Finally, we also compare against \baselinename{BOPE}, i.e., Bayesian Optimization with Preference Exploration~\citep{lin2022preference} in \Cref{sec:power-ablation}, which decomposes the two stages of preference exploration and experimentation.

\paragraph{Metrics.}  
Since we optimize over the preferences of LLM-proxy stakeholders, we do not actually have access to the ground truth preference function.
Instead, for each baseline and case study, we handcraft a deterministic \textit{preference score} function $h: y\to z$  representing the ground truth preference over objectives given the observables. 
The function is designed to be proportional to $f_{\text{subj}}$. 
To validate the correctness of the preference score function, we utilize TrueSkill Bayesian skill ranking~\citep{herbrich2006trueskill} to predict scores for each evaluation point. We also use other task specific metrics for evaluating the quality of generated scenarios, such as measuring distribution overlap with real data (\Cref{sec:travel-mode}), estimating coverage in feature space to measure the degree of exploration (\Cref{subsec:fire}), and qualitative visualizations (\Cref{subsec:power}).

\subsection{Power Grid Resource Allocation}\label{subsec:power}

We first study the ethical impact of using different Distributed Energy Resource (DER) deployment strategies with varying reactive power limits on the Power Grid Allocation IEEE 5-bus and IEEE 30-bus networks~\citep{luo2024ethical}, denoted \taskname{5-Bus} and \taskname{30-Bus} for convenience.

\paragraph{Scenario Description.} 
A scenario is parameterized by $x \coloneqq [l, r] \in \mathcal{X}$, where $l\in \{0, 1\}^{20}$ is a binary vector indicating if a certain location has DER deployment, and $r\in \mathbb{R}_{+}^{20}$ specifies the reactive power limits. 
This is a challenging problem for ethical evaluation due to multifaceted ethical concerns in distribution of resource arising from various stakeholders~\citep{luo2024ethical,NARUC_ADER_Fundamentals_2024}.
\begin{figure}
     \centering
     \begin{subfigure}[b]{\textwidth}
         \centering
         \includegraphics[width=0.8\textwidth]{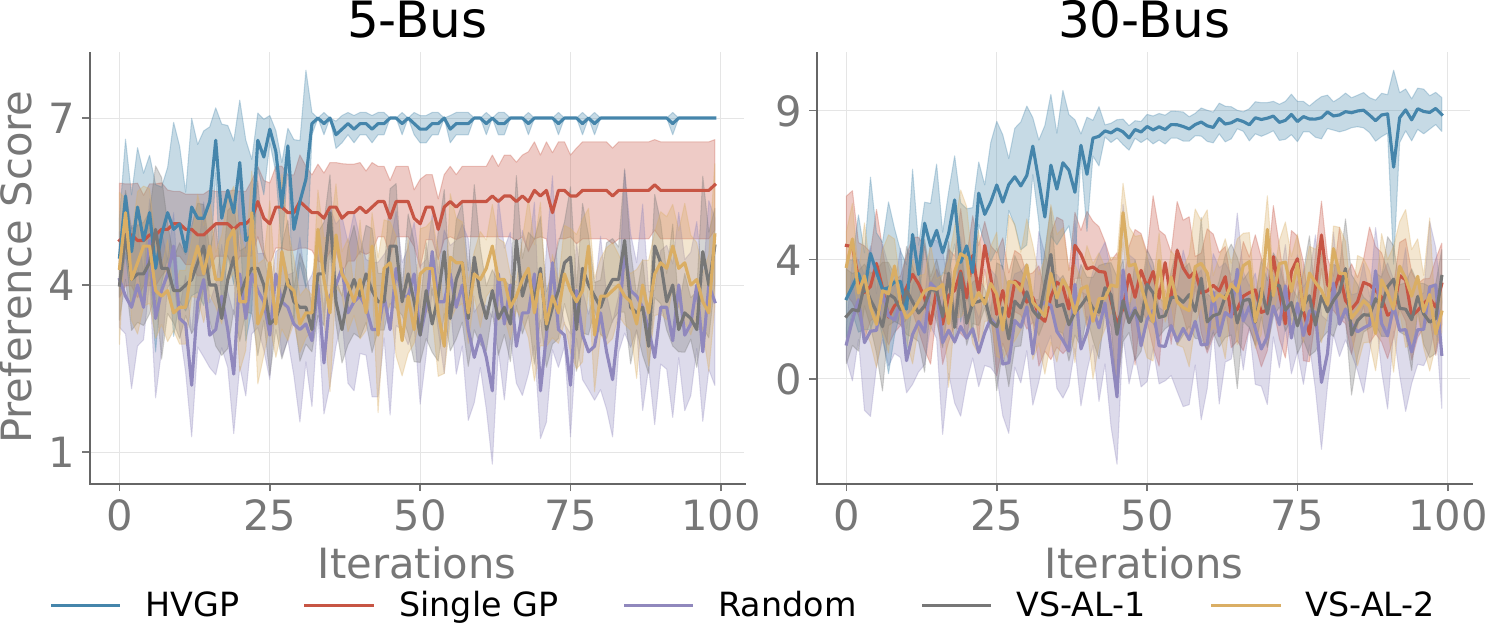}
     \end{subfigure}
     \caption{\textbf{Power Grid Allocation Preference Scores.} Preference scores baseline comparison for \taskname{5-Bus} (left) and \taskname{30-Bus} (right).}
      \label{fig:opf_main_pref}
\end{figure}

\paragraph{Observables.} 
Given scenario $x$, the resulting observables vector $y \in \mathbb{R}^4$ has four components.
The voltage \emph{Fairness} ($y^1$) measures the uniformity of voltage distribution across all buses. 
The total \emph{Cost} ($y^2$) combines the expenses of installing DER units and reactive power provision.
The \emph{Priority} ($y^3$) area coverage measures how well each design serves under-served or rural buses.
Finally, the \emph{Resilience} ($y^4$) assesses the network’s ability to maintain voltages above a specified threshold. 
We provide more details, including formulas, in ~\Cref{appendix:obj-opf}. 

\paragraph{Evaluation Method.}
In our prompt, we ask the LLM to prioritize Priority, followed by Cost, and ignore all other dimensions (prompt example in \Cref{appendix:eval-opf}).
Since we do not have access to ground truth evaluation scores, we approximate the LLM's preference function with a preference score function $\tilde{h}(y) \coloneqq  [0, -0.5, 1, 0]y$.

\paragraph{Results}
We evaluate on \fbus{} and \tbus{} and observe that our proposed \ouralgoname{HVGP} achieves a higher preference score than all other baselines.
Although \baselinename{Single GP} can do better than \baselinename{Random} on \fbus{}, it cannot solve \tbus{} since the design parameter dimensions grows to $40$, making it hard to efficiently explore the space.
On the contrary, \ouralgoname{HVGP} can mitigate this through the hierarchical structure, which reduces the complexity of mapping from objectives to subjective assessment, followed by our acquisition strategy, which prioritizes targeted exploration of objective space through the second mutual information term (MI2). 

Both \baselinename{VS-AL-1} and \baselinename{VS-AL-2} cannot solve either tasks, which we hypothesize is because of the inaccurate modeling choice in \baselinename{VS-AL-1} due to learning a linear decision boundary. Similar reasoning as \baselinename{Single GP} can be used to explain the inefficiency of 
\baselinename{VS-AL-2} due to its direct modeling of a complex decision space.

\begin{figure}
     \centering
     \begin{subfigure}[b]{\textwidth}
         \centering
         \includegraphics[width=0.8\textwidth]{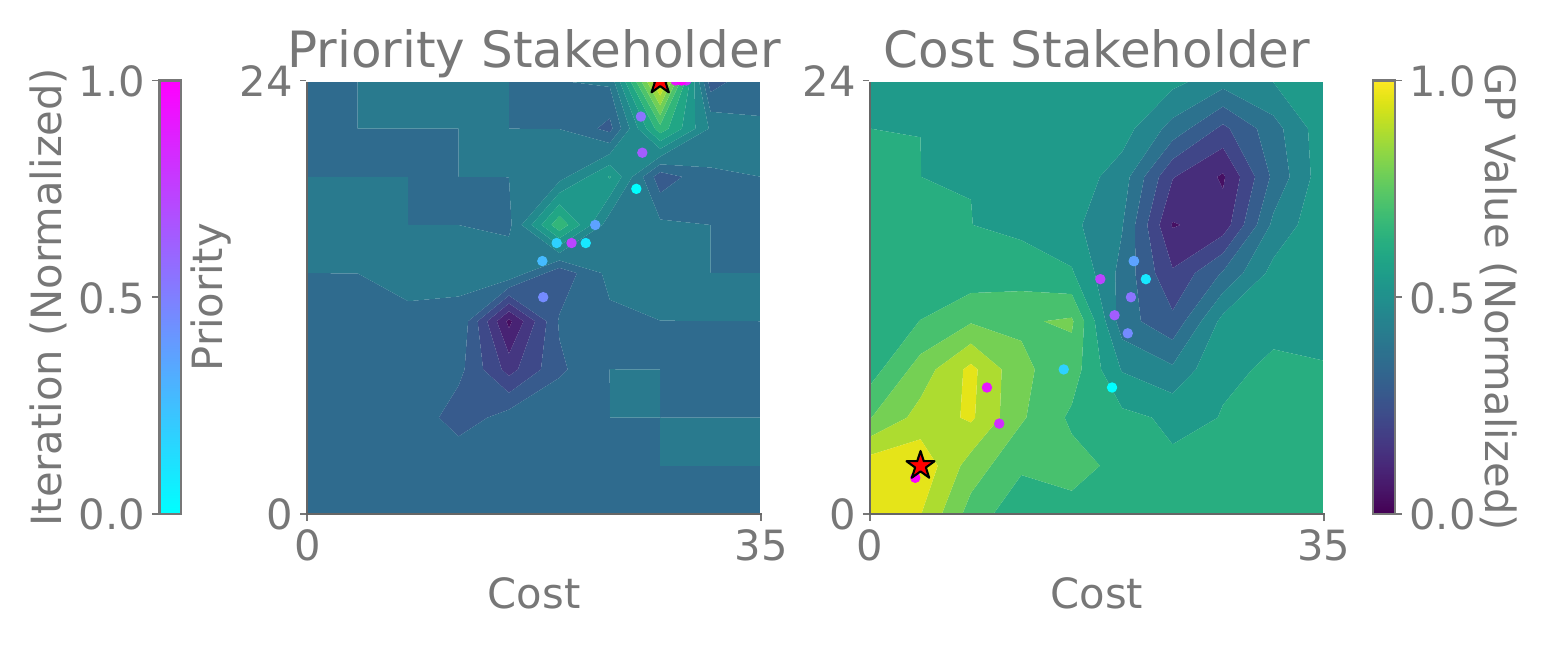}
     \end{subfigure}
     \caption{\textbf{Bus-30 Different Stakeholder Groups.} We show that our learned preference GP is able to adapt to the needs of different potential stakeholder groups. The plots show data point for optimum preference score (shown in red) projected on objective space, with contours of predicted preference score for Stakeholder A (left) and B (right), with optimum value data point.}
      \label{fig:OPF-cost-contour}
\end{figure}

\subsection{Fire Rescue}\label{subsec:fire}

Next, we consider an autonomous drone navigation scenario for fire extinguishing in a semi-urban setting, as motivated by discussions on ethical concerns in rescue robotics~\citep{battistuzzi2021ethical}. 

\paragraph{Scenario description.} 
A \taskname{Fire Rescue} scenario is parameterized by $x \in [0, 1]^{30}$, with only $9$ dimensions relevant to scenario design, controlling the placement of assets such as a museum, a gas station, a food court, and residential blocks with tree covers of variable density. The remaining dimensions are uncorrelated to the objectives.
In ~\Cref{fig:SAR-env-example} of ~\Cref{appendix:fire_rescue}), we give examples of three scenario visualizations.
The goal of the autonomous system is to search the area for fire and, based on its uncertainty, decide whether to continue exploring or spray retardant.


\paragraph{Observables.} 
Given scenario $x$, the observables vector $y = [y^1, y^2,y^3] \in \mathbb{R}^3$ quantifies the cumulative potential \emph{Chemical Damage} caused by deciding to spraying the retardant ($y^1$),
cumulative potential \emph{Fire Damage} caused by fire due to deciding not to spray the retardant ($y^2$), and \emph{Spread factor}($y^3$), measuring risk of firespread due to proximity of assets.

\paragraph{Evaluation Method.}
In our prompt, we ask the LLM to prioritize high Chemical damage and high Spread factor (prompt example in \Cref{appendix:fire_rescue}). We use preference score function $\tilde{h}(y) \coloneqq  [1,0,1]y$ and coverage score function $\tilde{c}(\textbf{x})$ that estimates cumulative standard deviation for $\textbf{x}=[x_1,\dots,x_n]$ for $n$ collected data points to measure the coverage of search space to sample novel scenarios.

\textbf{Results.}
We observe that \ouralgoname{HVGP} achieves higher preference score than all baselines, with a similar explanation as provided in the results of \Cref{subsec:power}.
We also observe that our acquisition strategy achieves higher preference score than two \ouralgoname{HVGP} variants: \baselinename{MI1+MI2}, and \baselinename{Pref}.
\baselinename{MI1+MI2} does not consider the preference term in the acquisition function, and \baselinename{Pref} does not consider the two mutual information terms.
The discrepancy in the preference scores is due to \baselinename{MI1+MI2}'s inefficient exploration in higher dimensions.
While \baselinename{Pref} performs better than \baselinename{MI1+MI2}, the complete acquisition strategy performs the best with the additional improvement from targeted exploration. Our method also shows higher coverage scores than baselines and ablations. We also provide an example of scenario generation using the learned model (~\Cref{appendix:scene-gen}). 


\begin{figure}
    \centering
    \includegraphics[width=0.45\linewidth]{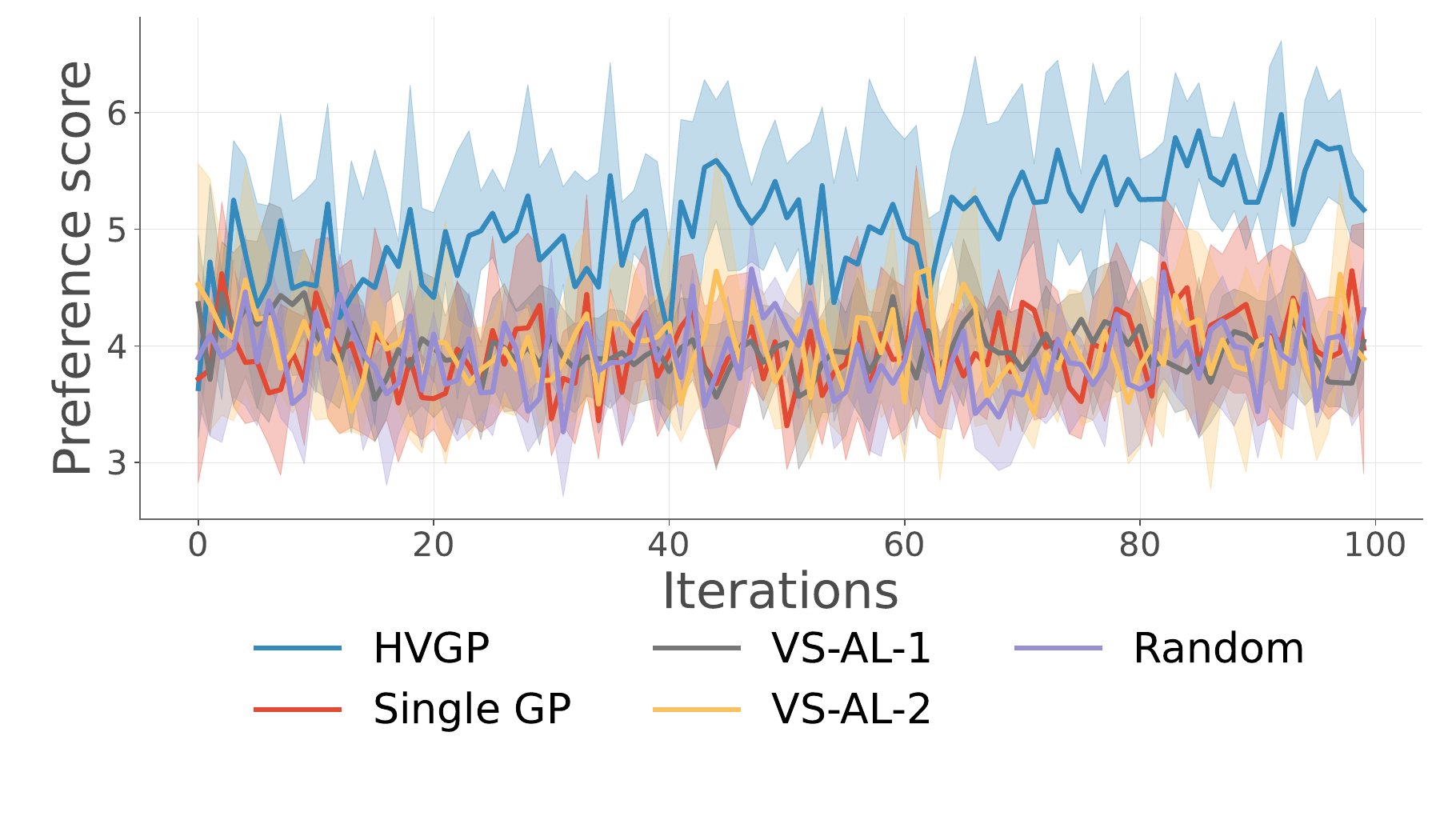}
    \includegraphics[width=0.45\linewidth]{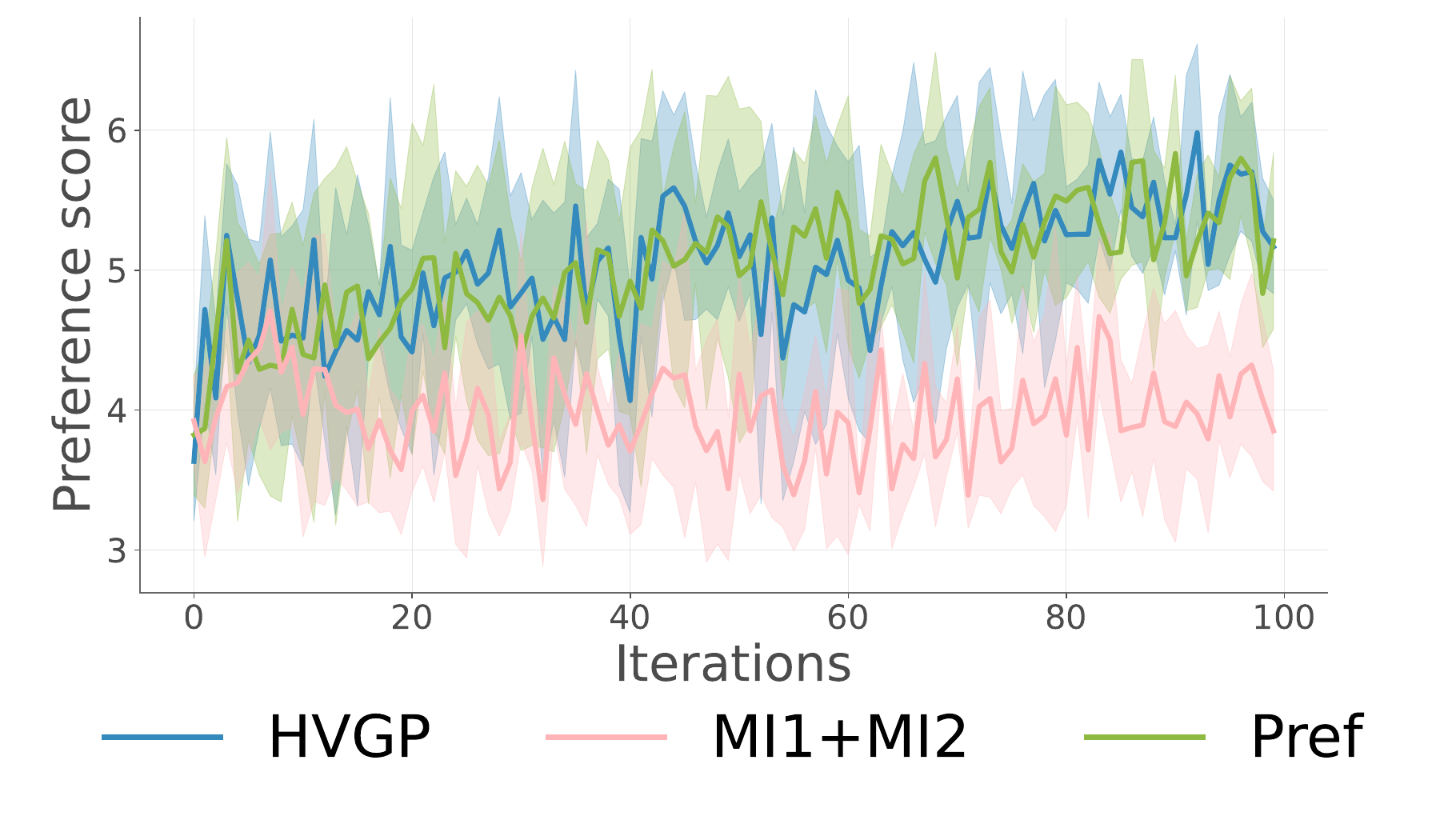}
    \caption{\textbf{\taskname{Fire Rescue} Preference Scores.} We report preferences scores for baseline comparisons (left) and acquisition strategy ablations (right). }
    \label{fig:fire_rescue_pref}
\end{figure}

\begin{figure}
    \centering
    \includegraphics[width=0.45\linewidth]{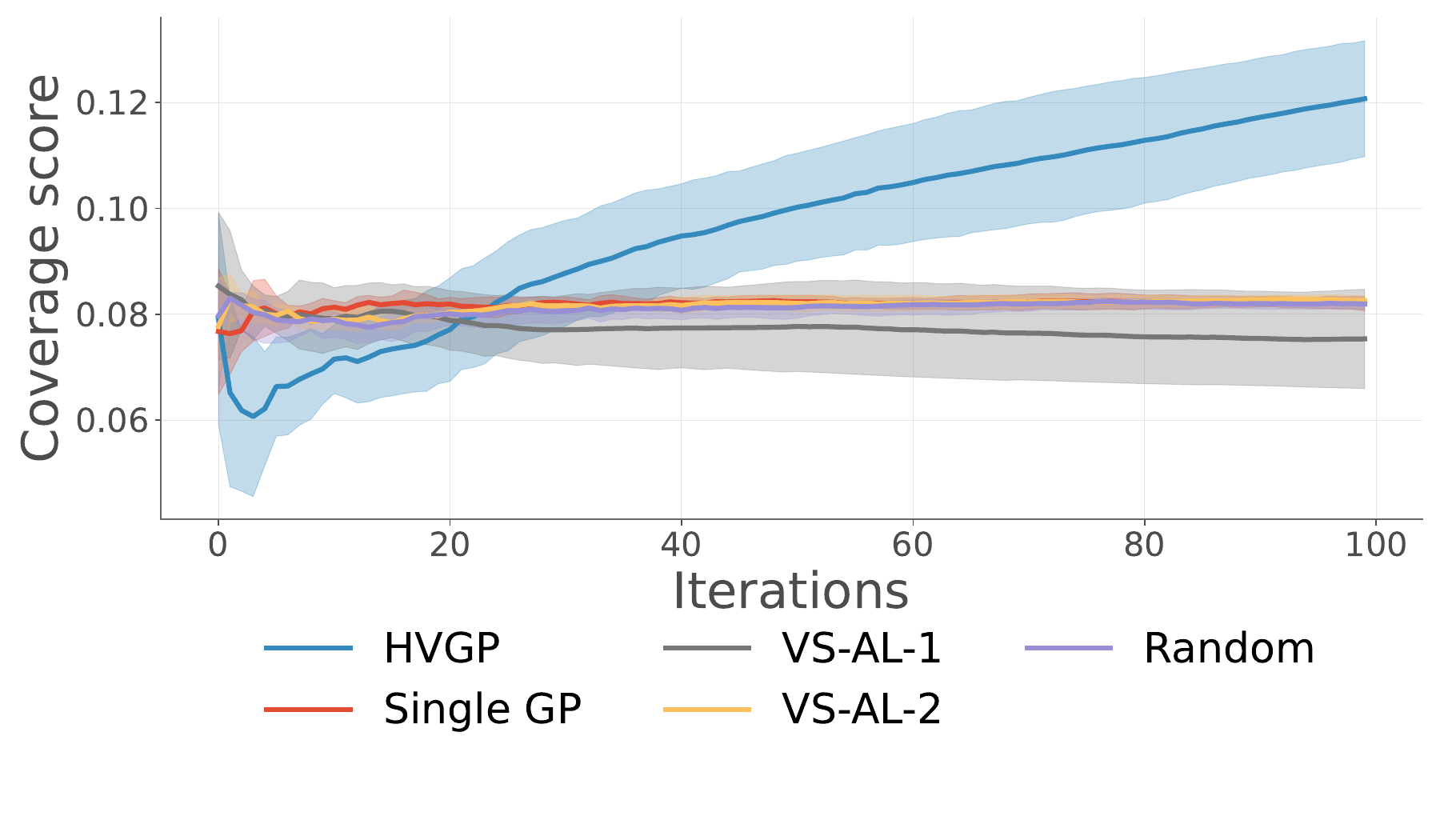}
    \includegraphics[width=0.45\linewidth]{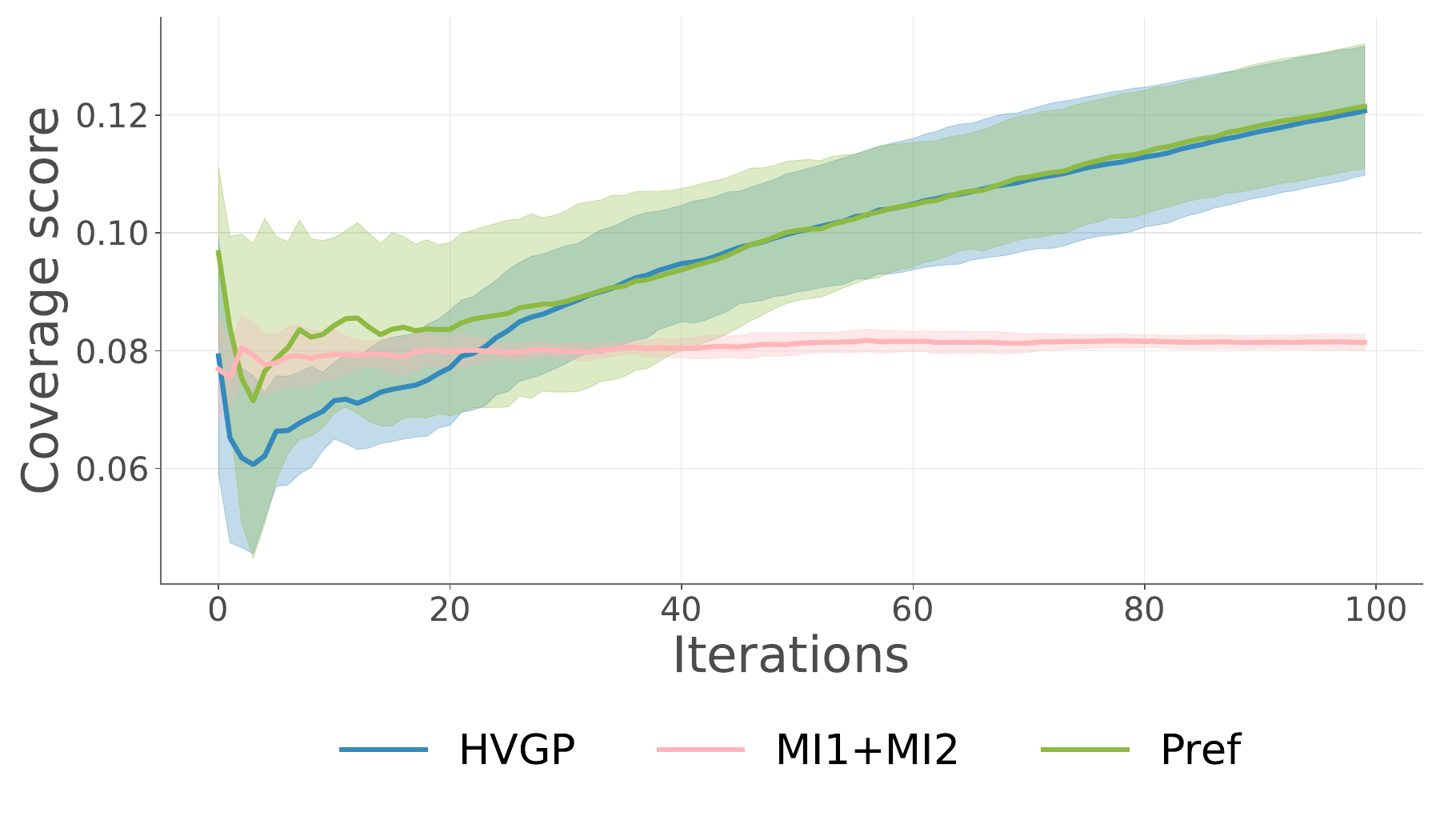}
    \caption{\textbf{\taskname{Fire Rescue} Coverage.} We report coverage scores for baseline comparisons (left) and acquisition strategy ablations (right).}
    \label{fig:fire_rescue_coverage}
\end{figure}

\subsection{Additional Results} \label{subsec:additional_results}

To better understanding how \AlgOurs{} works, we perform several ablation studies in a question-and-answer format.

\paragraph{(Q1) What conditions enable \AlgOurs{} to perform well?}
We observe that our method performs best when the design parameter space is large. 
In lower-dimensional settings such as \taskname{5-Bus}, \baselinename{Single GP} performs well but is still suboptimal compared to our method (~\Cref{fig:opf_main_pref} (left)).
However, in higher-dimensional cases such as \taskname{30-Bus} and \taskname{Fire Rescue}, \ouralgoname{HVGP} outperforms \baselinename{Single GP} by exploring test cases that highly align with the subjective criteria. We also observe superior performance of our method compared to ablations of \baselinename{BOPE} in \Cref{sec:power-ablation}, which shows that in limited sample settings, joint learning of objective and subjective models improves sample efficiency.
\begin{wrapfigure}[12]{r}{0.40\linewidth}
  \centering
  \includegraphics[width=\linewidth]{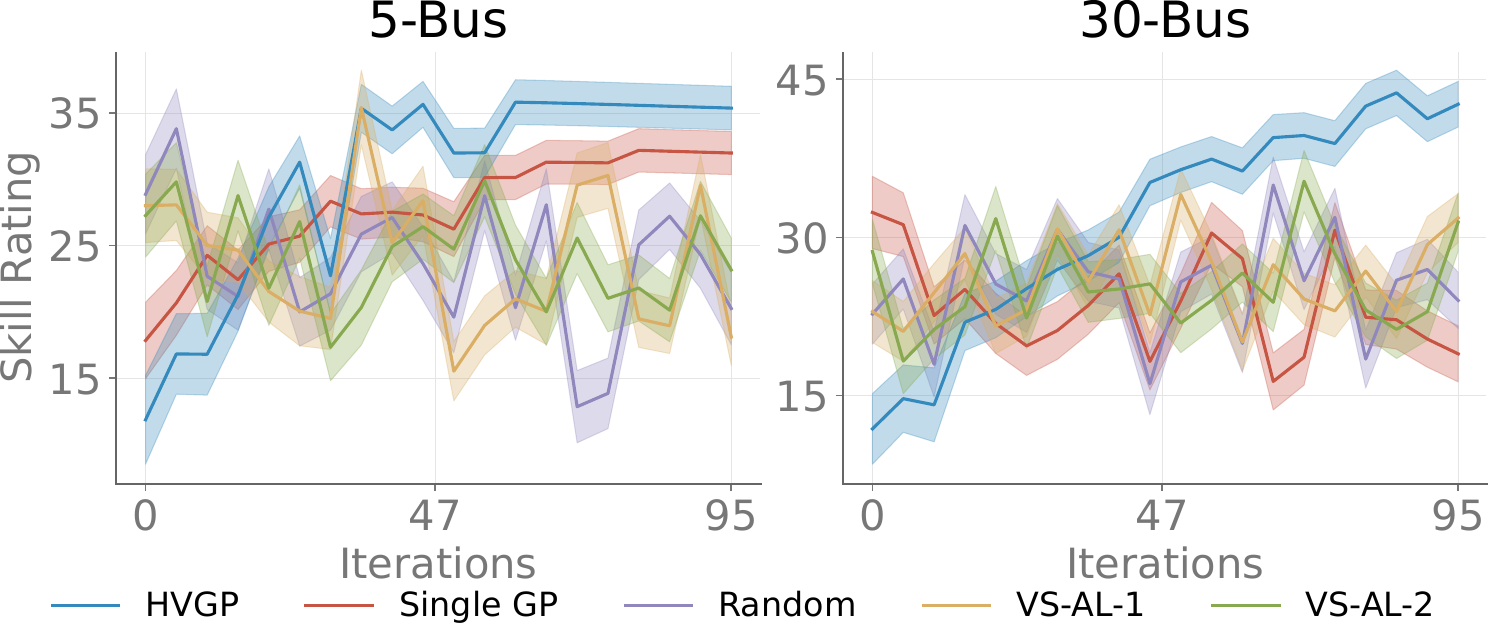}
  \caption{\textbf{Power Grid Allocation Skill Ratings.}
    Skill ratings computed using a Bayesian skill rating algorithm~\citep{herbrich2006trueskill}.}
  \label{fig:opf_main_bi}
\end{wrapfigure}
\noindent 

\vspace{-1em}
\paragraph{(Q2) Do our handcrafted preference score functions well approximate the LLM's preference function?}

It does. In \Cref{fig:opf_main_bi}, we treat each sampled observable as a player in a free-for-all game, and use the TrueSkill Bayesian skill rating system~\citep{herbrich2006trueskill} to evaluate their individual skill ratings (evaluation details in \Cref{subsubsec:trueskill_alg}).
This can be a more accurate evaluation method than our proposed preference scores since it does not assume the LLM's optimization objective form.
However, this evaluation process can be extremely expensive and the skill ratings are not directly comparable across seeds and baselines.
We observe that for both \taskname{5-Bus} and \taskname{30-Bus}, the trends roughly match the trends seen from the heuristic preference scores in \Cref{fig:opf_main_pref}.

\paragraph{(Q3) How does acquisition strategy support sample efficiency? }
In the acquisition ablation studies, we observe that our complete acquisition strategy consistently performs well. In \taskname{Fire Rescue} (right of ~\Cref{fig:fire_rescue_pref}), while preferential optimization is crucial, mutual information enables efficient exploration, which is essential in a high-dimensional search space with low volume optima, leading to incremental improvement over Pref. 
This validates our idea of balancing exploration and exploitation with the acquisition strategy.  

\paragraph{(Q4) How well does our model adapt to different stakeholders?}

 In power resource allocation, we consider an ablation with two stakeholders. Stakeholder A and B care mainly about high priority, and low cost respectively. ~\Cref{fig:OPF-cost-contour} shows the optimum value scenario for Stakeholder A has a high priority score, and high cost, whereas for Stakeholder B, the optimum corresponds to low cost and low priority. This shows that the sampling procedure accurately accounts for the stakeholder-specific criteria, resulting in different explorations, and stakeholder-specific test cases. We also extend this analysis to a more challenging, multi-stakeholder setting for the Power Grid case study (details in~\Cref{sec:multi-stakeholder}), \Cref{fig:ms-criteria} and observe similar trends.

 \paragraph{(Q5) How robust is our approach to LLM specifications?}

We conduct ablation studies for the Fire Rescue case study for LLM parameters, varying temperature, prompt and model (\Cref{sec:llm-ablation}), and observe robustness to evaluator perturbations. We attribute this to pairwise elicitation, that eliminates uncertainty from self-inconsistency~\citep{bouwer2024comparative,hoeijmakers2024subjective}.

\paragraph{(Q6) How suitable is our framework for real-world ethical alignment?}
In \Cref{sec:travel-mode}, we apply our framework to extract latent objectives influencing travel mode choices using the TravelMode dataset~\citep{greene2003econometric}.  This case study shows that our framework can be effectively applied to learn underlying trends in objective landscape (\Cref{fig:travel-mode}).

\section{Limitations}\label{sec:limitations}
While the SEED-SET framework effectively mitigates common challenges associated with ethical assessment, certain limitations remain, along with promising future directions.


\vspace{-0.5em}
 \paragraph{Scalability for Extremely Large Datasets.} Using sparse variational GPs reduces complexity from $O(N^3)$ to $O(NM^2)$ with $M$ inducing points, enabling SEED-SET to handle tens of thousands of observations. Scaling to hundreds of thousands or more remains challenging, which future work could address via stochastic variational inference (SVI). 

 The current model uses a stationary kernel (e.g., RBF), assuming covariance depends only on relative distance. This can be too restrictive for systems with varying regimes. To relax this, SEED-SET can be extended with non-stationary kernels (e.g., spectral mixture, input-warped) or deep GPs that warp inputs through neural layers.
Our model requires complete knowledge of objective metrics a priori. While this is a reasonable assumption that mimics real-world preference elicitation, in the lack of complete list of objectives, the testing process can have inaccuracies.



Using LLMs as ethical proxies also risks sensitivity to prompts and context, so ongoing alignment checks or fine-tuning are needed to keep their judgments in sync with human values.
Still, when no ground truth exists, preference data is a practical surrogate, and VGPs with Bayesian design offer a sample-efficient solution that would be far cheaper than training an LLM from scratch on preferences.

\vspace{-0.5em}
\section{Conclusion}
We presented SEED-SET, a scalable framework for ethical evaluation of autonomous systems that combines objective system metrics with subjective stakeholder judgments through a hierarchical variational Bayesian model. By separating measurable factors from user preferences and guiding exploration via a principled acquisition strategy, SEED-SET enables efficient and interpretable evaluation of ethical trade-offs. The integration of large language models as proxy evaluators further reduces human burden while maintaining value alignment. Experiments across domains demonstrate SEED-SET’s effectiveness, with future work aimed at extending to multi-agent settings and real-time applications.


\section{Ethics statement}
In this work, we propose an automated evaluation tool for ethical assessment of autonomous systems. Our work assumes a provided ethical criteria for evaluation, and associated cost functions.  We make no recommendations or comments on the correctness of any ethical criteria in this work, and mainly leverage existing instances of ethical evaluation for validation of our pipeline. The paper does not involve crowdsourcing or research with human subjects.

\section{Reproducibility Statement}
All simulations for Fire rescue, Power resource grid allocation and Optimal Routing were conducted on a Linux workstation with Ubuntu 22.04 LTS equipped with an Intel 13th Gen Core i7-13700KF CPU (16 cores, 24 threads, up to 5.4 GHz) and an NVIDIA GeForce RTX 4090 GPU (24 GB VRAM). The Bayesian Experimental Design (BED) loops were implemented using wrapper BoTorch~\citep{balandat2020botorch} and GPyTorch~\citep{gardner2018gpytorch} libraries. The compute requirements were consistent with standard usage of these libraries and did not require additional specialized hardware beyond what was used for \textsf{Webots} simulation. Implementation specific details of both the simulations are provided in \Cref{appendix:opf} and \Cref{appendix:fire_rescue}. Examples of LLM prompts used in the generation of results reported in the paper are also provided in the Appendix. 

\section{Acknowledgement}

This research was developed with funding from the Defense Advanced Research Projects Agency (DARPA) {with Distribution Statement "A" (Approved for Public Release, Distribution Unlimited)} . The views, opinions and/or findings expressed are those of the author and should not be
interpreted as representing the official views or policies of DARPA or the U.S. Government.

 We thank Jonathan Becker, Ashley Wiren, and Mitul Saha for their valuable discussions and support throughout the project.

\newpage
\bibliography{refs}

@article{hagendorff2020ethics,
  title={The ethics of AI ethics: An evaluation of guidelines},
  author={Hagendorff, Thilo},
  journal={Minds and machines},
  volume={30},
  number={1},
  pages={99--120},
  year={2020},
  publisher={Springer}
}

@misc{ai-rmf,
  author        = {Elham Tabassi},
  title         = {Artificial intelligence risk management framework (AI RMF 1.0)},
  year          = {2023},
  month         = {January},
  howpublished  = {\url{https://tsapps.nist.gov/publication/get_pdf.cfm?pub_id=936225}},
  note          = {National Institute of Standards and Technology, Gaithersburg, MD. \doi{10.6028/NIST.AI.100-1}}
}

@article{chance2023assessing,
  title={Assessing trustworthiness of autonomous systems},
  author={Chance, Gregory and Abeywickrama, Dhaminda B and LeClair, Beckett and Kerr, Owen and Eder, Kerstin},
  journal={arXiv preprint arXiv:2305.03411},
  year={2023}
}

@article{liu2024aligning,
  title={Aligning with human judgement: The role of pairwise preference in large language model evaluators},
  author={Liu, Yinhong and Zhou, Han and Guo, Zhijiang and Shareghi, Ehsan and Vuli{\'c}, Ivan and Korhonen, Anna and Collier, Nigel},
  journal={arXiv preprint arXiv:2403.16950},
  year={2024}
}

@article{michel2004cyberbotics,
  title={Cyberbotics ltd. webots™: professional mobile robot simulation},
  author={Michel, Olivier},
  journal={International Journal of Advanced Robotic Systems},
  volume={1},
  number={1},
  pages={5},
  year={2004},
  publisher={SAGE Publications Sage UK: London, England}
}

@article{IEEE-P7001,
    author = {Winfield, A. F. T. and Booth, S. and Dennis, L. A. and Egawa, T. and Hastie, H. and Jacobs, N. and Muttram, R. I. and Olszewska, J. I. and Rajabiyazdi, F. and Theodorou, A. and Underwood, M. A. and Wortham, R. H. and Watson, E.},
    title = {IEEE P7001: A Proposed Standard on Transparency},
    year = {2021},
    journal = {Frontiers in robotics and AI},
    volume={8},
    pages={665729},
}

@misc{ISO/PAS-8800,
  title = {ISO/PAS-8800 Road vehicles — Safety and artificial
intelligence},
  year = {2024},
}

@article{hoeijmakers2024subjective,
  title={How subjective CT image quality assessment becomes surprisingly reliable: pairwise comparisons instead of Likert scale},
  author={Hoeijmakers, Eva JI and Martens, Bibi and Hendriks, Babs MF and Mihl, Casper and Miclea, Razvan L and Backes, Walter H and Wildberger, Joachim E and Zijta, Frank M and Gietema, Hester A and Nelemans, Patricia J and others},
  journal={European Radiology},
  volume={34},
  number={7},
  pages={4494--4503},
  year={2024},
  publisher={Springer}
}

@article{bouwer2024comparative,
  title={Comparative approaches to the assessment of writing: Reliability and validity of benchmark rating and comparative judgement},
  author={Bouwer, Renske and Lesterhuis, Marije and De Smedt, Fien and Van Keer, Hilde and De Maeyer, Sven},
  journal={Journal of Writing Research},
  volume={15},
  number={3},
  pages={498--518},
  year={2024},
  publisher={University of Antwerp}
}

@book{martens2016transport,
  title={Transport justice: Designing fair transportation systems},
  author={Martens, Karel},
  year={2016},
  publisher={Routledge}
}

@article{pereira2017distributive,
  title={Distributive justice and equity in transportation},
  author={Pereira, Rafael HM and Schwanen, Tim and Banister, David},
  journal={Transport reviews},
  volume={37},
  number={2},
  pages={170--191},
  year={2017},
  publisher={Taylor \& Francis}
}

@book{greene2003econometric,
  title        = {Econometric Analysis},
  author       = {Greene, William H.},
  edition      = {5},
  year         = {2003},
  publisher    = {Prentice Hall},
  address      = {Upper Saddle River, NJ}
}

@inproceedings{astudillo2019bayesian,
  title={Bayesian optimization of composite functions},
  author={Astudillo, Raul and Frazier, Peter},
  booktitle={International Conference on Machine Learning},
  pages={354--363},
  year={2019},
  organization={PMLR}
}

@inproceedings{lin2022preference,
  title={Preference exploration for efficient bayesian optimization with multiple outcomes},
  author={Lin, Zhiyuan Jerry and Astudillo, Raul and Frazier, Peter and Bakshy, Eytan},
  booktitle={International Conference on Artificial Intelligence and Statistics},
  pages={4235--4258},
  year={2022},
  organization={PMLR}
}

@article{rainforth2024modern,
  title={Modern Bayesian experimental design},
  author={Rainforth, Tom and Foster, Adam and Ivanova, Desi R and Bickford Smith, Freddie},
  journal={Statistical Science},
  volume={39},
  number={1},
  pages={100--114},
  year={2024},
  publisher={Institute of Mathematical Statistics}
}

@article{shao2023eliciting,
  title={Eliciting user preferences for personalized multi-objective decision making through comparative feedback},
  author={Shao, Han and Cohen, Lee and Blum, Avrim and Mansour, Yishay and Saha, Aadirupa and Walter, Matthew},
  journal={Advances in Neural Information Processing Systems},
  volume={36},
  pages={12192--12221},
  year={2023}
}

@article{Christiano2017DeepRL,
  title={Deep Reinforcement Learning from Human Preferences},
  author={Paul Francis Christiano and Jan Leike and Tom B. Brown and Miljan Martic and Shane Legg and Dario Amodei},
  journal={ArXiv},
  year={2017},
  volume={abs/1706.03741},
  url={https://api.semanticscholar.org/CorpusID:4787508}
}

@techreport{NARUC_ADER_Fundamentals_2024,
  title        = {Aggregated Distributed Energy Resources in 2024: The Fundamentals},
  author       = {Stephanie Bieler and Cara Goldenberg and Avery McEvoy and Katerina Stephan and Alex Walmsley},
  institution  = {RMI, under contract with NARUC},
  year         = {2024},
  month        = jul,
  note         = {Prepared as part of the NARUC–NASEO DER Integration and Compensation Initiative, DOE Award No.\ DE-OE0000925},
  url          = {https://connectedcommunities.lbl.gov/sites/default/files/2024-07/NARUC_ADER_Fundamentals_Interactive.pdf}
}

@article{tran2015variational,
  title={The variational Gaussian process},
  author={Tran, Dustin and Ranganath, Rajesh and Blei, David M},
  journal={arXiv preprint arXiv:1511.06499},
  year={2015}
}

@article{gao2024ethical,
  title={Ethical alignment decision-making for connected autonomous vehicle in traffic dilemmas via reinforcement learning from human feedback},
  author={Gao, Xin and Luan, Tian and Li, Xueyuan and Liu, Qi and Ma, Zhaoyang and Meng, Xiaoqiang and Li, Zirui},
  journal={IEEE Internet of Things Journal},
  year={2024},
  publisher={IEEE}
}

@article{dennis2016formal,
  title={Formal verification of ethical choices in autonomous systems},
  author={Dennis, Louise and Fisher, Michael and Slavkovik, Marija and Webster, Matt},
  journal={Robotics and Autonomous Systems},
  volume={77},
  pages={1--14},
  year={2016},
  publisher={Elsevier}
}

@inproceedings{reuel2022using,
  title={Using Adaptive Stress Testing to Identify Paths to Ethical Dilemmas in Autonomous Systems.},
  author={Reuel, Ann-Katrin and Koren, Mark and Corso, Anthony and Kochenderfer, Mykel J},
  booktitle={SafeAI@ AAAI},
  year={2022}
}

@inproceedings{keswani2024pros,
  title={On the Pros and Cons of Active Learning for Moral Preference Elicitation},
  author={Keswani, Vijay and Conitzer, Vincent and Heidari, Hoda and Borg, Jana Schaich and Sinnott-Armstrong, Walter},
  booktitle={Proceedings of the AAAI/ACM Conference on AI, Ethics, and Society},
  volume={7},
  pages={711--723},
  year={2024}
}

@article{tarsney2024moral,
  title={Moral Decision-Making Under Uncertainty},
  author={Tarsney, Christian and Thomas, Teruji and MacAskill, William},
  year={2024}
}

@article{sovacool2016energy,
  title={Energy decisions reframed as justice and ethical concerns},
  author={Sovacool, Benjamin K and Heffron, Raphael J and McCauley, Darren and Goldthau, Andreas},
  journal={Nature Energy},
  volume={1},
  number={5},
  pages={1--6},
  year={2016},
  publisher={Nature Publishing Group}
}

@article{bhattacharya2024trust,
  title={Trust \& fair resource allocation in community energy systems},
  author={Bhattacharya, Sanjukta and Rousis, Anastasios Oulis and Bourazeri, Aikaterini},
  journal={IEEE Access},
  volume={12},
  pages={11157--11169},
  year={2024},
  publisher={IEEE}
}

@article{fahmin2024eco,
  title={Eco-Friendly Route Planning Algorithms: Taxonomies, Literature Review and Future Directions},
  author={Fahmin, Ahmed and Cheema, Muhammad Aamir and Eunus Ali, Mohammed and Nadjaran Toosi, Adel and Lu, Hua and Li, Huan and Taniar, David and A. Rakha, Hesham and Shen, Bojie},
  journal={ACM Computing Surveys},
  volume={57},
  number={1},
  pages={1--42},
  year={2024},
  publisher={ACM New York, NY}
}

@article{chitikena2023robotics,
  title={Robotics in search and rescue (sar) operations: An ethical and design perspective framework for response phase},
  author={Chitikena, Hareesh and Sanfilippo, Filippo and Ma, Shugen},
  journal={Applied Sciences},
  volume={13},
  number={3},
  pages={1800},
  year={2023},
  publisher={MDPI}
}

@article{Chaloner1995BayesianReview,
    title = {{Bayesian experimental design: A review}},
    year = {1995},
    journal = {Statistical science},
    author = {Chaloner, Kathryn and Verdinelli, Isabella},
    pages = {273--304}
}

@misc{huang2025acceleratedpreferenceelicitationllmbased,
      title={Accelerated Preference Elicitation with LLM-Based Proxies}, 
      author={David Huang and Francisco Marmolejo-Cossío and Edwin Lock and David Parkes},
      year={2025},
      eprint={2501.14625},
      archivePrefix={arXiv},
      primaryClass={cs.GT},
      url={https://arxiv.org/abs/2501.14625}, 
}

@inproceedings{Titsias2009VariationalProcesses,
    title = {{Variational learning of inducing variables in sparse Gaussian processes}},
    year = {2009},
    booktitle = {Artificial intelligence and statistics},
    author = {Titsias, Michalis},
    pages = {567--574}
}

@inproceedings{Chu2005PreferenceProcesses,
    title = {{Preference learning with Gaussian processes}},
    year = {2005},
    booktitle = {Proceedings of the 22nd international conference on Machine learning},
    author = {Chu, Wei and Ghahramani, Zoubin},
    pages = {137--144}
}

@article{mehrabi2021fairness,
  title={A survey on bias and fairness in machine learning},
  author={Mehrabi, Ninareh and Morstatter, Fred and Saxena, Nripsuta and Lerman, Kristina and Galstyan, Aram},
  journal={arXiv preprint arXiv:1908.09635},
  year={2021}
}

@article{xu2020quantification,
  title={Towards the Quantification of Safety Risks in Deep Neural Networks},
  author={Xu, Peipei and Ruan, Wenjie and Huang, Xiaowei},
  journal={arXiv preprint arXiv:2009.06114},
  year={2020}
}

@article{sensoy2024riskaware,
  title={Risk-aware Classification via Uncertainty Quantification},
  author={Sensoy, Murat and Kaplan, Lance M. and Julier, Simon and Saleki, Maryam and Cerutti, Federico},
  journal={arXiv preprint arXiv:2412.03391},
  year={2024},
  howpublished={\url{https://arxiv.org/abs/2412.03391}}
}

@article{hernandez2023toolkit,
  title={A toolkit of dilemmas: Beyond debiasing and fairness formulas for responsible AI/ML},
  author={Domínguez Hernández, Andrés and Galanos, Vassilis},
  journal={arXiv preprint arXiv:2303.01930},
  year={2023}
}

@article{weidinger2023sociotechnical,
  title={Sociotechnical safety evaluation of generative AI systems},
  author={Weidinger, Laura and Rauh, Maribeth and Marchal, Nahema and Manzini, Arianna and Hendricks, Lisa Anne and Mateos-Garcia, Juan and Bergman, Stevie and Kay, Jackie and Griffin, Conor and Bariach, Ben and others},
  journal={arXiv preprint arXiv:2310.06682},
  year={2023}
}

@article{jiang2021delphi,
  title={Can Machines Learn Morality? The Delphi Experiment},
  author={Jiang, Liwei and Hwang, Jena D. and Bhagavatula, Chandra and Le Bras, Ronan and Liang, Jenny and Dodge, Jesse and Sakaguchi, Keisuke and Forbes, Maxwell and Borchardt, Jon and Gabriel, Saadia and Tsvetkov, Yulia and Etzioni, Oren and Sap, Maarten and Rini, Regina and Choi, Yejin},
  journal={arXiv preprint arXiv:2110.07574},
  year={2021}
}

@article{ziegler2022selfrewarding,
  title={Self-rewarding language models},
  author={Ziegler, Daniel and Ouyang, Long and Wu, Jeffrey and Lowe, Ryan and Stiennon, Nisan and Gray, John and Schulman, John and Christiano, Paul},
  journal={arXiv preprint arXiv:2210.01152},
  year={2022}
}

@misc{oecd2019principles,
  author        = {{Organisation for Economic Co-operation and Development}},
  title         = {OECD principles on artificial intelligence},
  year          = {2019},
  howpublished  = {\url{https://www.oecd.org/en/topics/sub-issues/ai-principles.html}}
}

@misc{ieee2019aligned,
  author        = {{IEEE Global Initiative}},
  title         = {Ethically aligned design: A vision for prioritizing human well-being with autonomous and intelligent systems},
  year          = {2019},
  howpublished  = {\url{https://sagroups.ieee.org/global-initiative/wp-content/uploads/sites/542/2023/01/ead1e.pdf}},
  publisher     = {IEEE Standards Association}
}

@article{cong2022energyequitygap,
  title        = {Unveiling hidden energy poverty using the energy equity gap},
  author       = {Cong, Shiyang and Nock, Destenie and Qiu, Yuchen L. and Xing, Bo},
  journal      = {Nature Communications},
  volume       = {13},
  number       = {1},
  pages        = {2456},
  year         = {2022},
  publisher    = {Nature},
  doi          = {10.1038/s41467-022-30146-5},
  howpublished = {\url{https://www.nature.com/articles/s41467-022-30146-5}}
}

@article{balandat2020botorch,
  title={BoTorch: A framework for efficient Monte-Carlo Bayesian optimization},
  author={Balandat, Maximilian and Karrer, Brian and Jiang, Daniel and Daulton, Samuel and Letham, Ben and Wilson, Andrew G and Bakshy, Eytan},
  journal={Advances in neural information processing systems},
  volume={33},
  pages={21524--21538},
  year={2020}
}

@article{gardner2018gpytorch,
  title={Gpytorch: Blackbox matrix-matrix gaussian process inference with gpu acceleration},
  author={Gardner, Jacob and Pleiss, Geoff and Weinberger, Kilian Q and Bindel, David and Wilson, Andrew G},
  journal={Advances in neural information processing systems},
  volume={31},
  year={2018}
}

@article{zaim2025large,
  title={Large-scale moral machine experiment on large language models},
  author={Zaim bin Ahmad, Muhammad Shahrul and Takemoto, Kazuhiro},
  journal={PloS one},
  volume={20},
  number={5},
  pages={e0322776},
  year={2025},
  publisher={Public Library of Science San Francisco, CA USA}
}

@article{amodei2016concrete,
  title={Concrete problems in AI safety},
  author={Amodei, Dario and Olah, Chris and Steinhardt, Jacob and Christiano, Paul and Schulman, John and Man{\'e}, Dan},
  journal={arXiv preprint arXiv:1606.06565},
  year={2016}
}

@article{jobin2019global,
  title={The global landscape of AI ethics guidelines},
  author={Jobin, Anna and Ienca, Marcello and Vayena, Effy},
  journal={Nature machine intelligence},
  volume={1},
  number={9},
  pages={389--399},
  year={2019},
  publisher={Nature Publishing Group UK London}
}

@article{mittelstadt2016ethics,
  title={The ethics of algorithms: Mapping the debate},
  author={Mittelstadt, Brent Daniel and Allo, Patrick and Taddeo, Mariarosaria and Wachter, Sandra and Floridi, Luciano},
  journal={Big Data \& Society},
  volume={3},
  number={2},
  pages={2053951716679679},
  year={2016},
  publisher={SAGE Publications Sage UK: London, England}
}

@article{maslej2025artificial,
  title={Artificial intelligence index report 2025},
  author={Maslej, Nestor and Fattorini, Loredana and Perrault, Raymond and Gil, Yolanda and Parli, Vanessa and Kariuki, Njenga and Capstick, Emily and Reuel, Anka and Brynjolfsson, Erik and Etchemendy, John and others},
  journal={arXiv preprint arXiv:2504.07139},
  year={2025}
}

@article{zeng2024air,
  title={Air-bench 2024: A safety benchmark based on risk categories from regulations and policies},
  author={Zeng, Yi and Yang, Yu and Zhou, Andy and Tan, Jeffrey Ziwei and Tu, Yuheng and Mai, Yifan and Klyman, Kevin and Pan, Minzhou and Jia, Ruoxi and Song, Dawn and others},
  journal={arXiv preprint arXiv:2407.17436},
  year={2024}
}

@article{weidinger2021ethical,
  title={Ethical and social risks of harm from language models},
  author={Weidinger, Laura and Mellor, John and Rauh, Maribeth and Griffin, Conor and Uesato, Jonathan and Huang, Po-Sen and Cheng, Myra and Glaese, Mia and Balle, Borja and Kasirzadeh, Atoosa and others},
  journal={arXiv preprint arXiv:2112.04359},
  year={2021}
}

@inproceedings{birhane2024dark,
  title={The dark side of dataset scaling: Evaluating racial classification in multimodal models},
  author={Birhane, Abeba and Dehdashtian, Sepehr and Prabhu, Vinay and Boddeti, Vishnu},
  booktitle={Proceedings of the 2024 ACM Conference on Fairness, Accountability, and Transparency},
  pages={1229--1244},
  year={2024}
}

@inproceedings{wang2025measuring,
  title={Measuring Machine Learning Harms from Stereotypes Requires Understanding Who Is Harmed by Which Errors in What Ways},
  author={Wang, Angelina and Bai, Xuechunzi and Barocas, Solon and Blodgett, Su Lin},
  booktitle={Proceedings of the 2025 ACM Conference on Fairness, Accountability, and Transparency},
  pages={746--762},
  year={2025}
}

@article{grabb2024risks,
  title={Risks from language models for automated mental healthcare: Ethics and structure for implementation},
  author={Grabb, Declan and Lamparth, Max and Vasan, Nina},
  journal={arXiv preprint arXiv:2406.11852},
  year={2024}
}

@article{salaudeen2025measurement,
  title={Measurement to Meaning: A Validity-Centered Framework for AI Evaluation},
  author={Salaudeen, Olawale and Reuel, Anka and Ahmed, Ahmed and Bedi, Suhana and Robertson, Zachary and Sundar, Sudharsan and Domingue, Ben and Wang, Angelina and Koyejo, Sanmi},
  journal={arXiv preprint arXiv:2505.10573},
  year={2025}
}

@article{reuel2024open,
  title={Open problems in technical ai governance},
  author={Reuel, Anka and Bucknall, Ben and Casper, Stephen and Fist, Tim and Soder, Lisa and Aarne, Onni and Hammond, Lewis and Ibrahim, Lujain and Chan, Alan and Wills, Peter and others},
  journal={arXiv preprint arXiv:2407.14981},
  year={2024}
}

@article{wallach2025position,
  title={Position: Evaluating generative ai systems is a social science measurement challenge},
  author={Wallach, Hanna and Desai, Meera and Cooper, A Feder and Wang, Angelina and Atalla, Chad and Barocas, Solon and Blodgett, Su Lin and Chouldechova, Alexandra and Corvi, Emily and Dow, P Alex and others},
  journal={arXiv preprint arXiv:2502.00561},
  year={2025}
}

@article{palka2023ai,
  title={AI, Consumers \& Psychological Harm},
  author={Pa{\l}ka, Przemys{\l}aw},
  journal={AI and Consumers,'Larry DiMatteo, Cristina Poncib{\`o}, Martin Hogg, Geraint Howells (Eds.), Cambridge University Press (2023/2024)},
  year={2023}
}

@article{noothigattu2019teaching,
  title={Teaching AI agents ethical values using reinforcement learning and policy orchestration},
  author={Noothigattu, Ritesh and Bouneffouf, Djallel and Mattei, Nicholas and Chandra, Rachita and Madan, Piyush and Varshney, Kush R and Campbell, Murray and Singh, Moninder and Rossi, Francesca},
  journal={IBM Journal of Research and Development},
  volume={63},
  number={4/5},
  pages={2--1},
  year={2019},
  publisher={IBM}
}

@article{luo2024ethical,
  title={Ethical Considerations in Smart Grid Optimization Promoting Energy Equity and Fairness},
  author={Luo, Dongwen and Zhong, Jiachen and Wang, Yiting and Pan, Weihao},
  year={2024},
  publisher={Preprints}
}

@article{battistuzzi2021ethical,
  title={Ethical concerns in rescue robotics: a scoping review},
  author={Battistuzzi, Linda and Recchiuto, Carmine Tommaso and Sgorbissa, Antonio},
  journal={Ethics and Information Technology},
  volume={23},
  number={4},
  pages={863--875},
  year={2021},
  publisher={Springer}
}

@inproceedings{herbrich2006trueskill,
author = {Herbrich, Ralf and Minka, Tom and Graepel, Thore},
title = {TrueSkill™: a Bayesian skill rating system},
year = {2006},
publisher = {MIT Press},
address = {Cambridge, MA, USA},
booktitle = {Proceedings of the 20th International Conference on Neural Information Processing Systems},
pages = {569–576},
numpages = {8},
location = {Canada},
series = {NIPS'06}
}

@misc{trueskill,
  title = {Trueskill: the video game rating system},
  howpublished = {\url{https://trueskill.org/}},
}
\bibliographystyle{iclr2026_conference}


\newpage
\appendix
\section{Literature Review}
\subsection{Additional Related Works}\label{sec:lit}

\textbf{Governance Approaches, for Responsible AI.} A wide range of governance frameworks, guidelines and standards have been proposed to guide the ethical development and deployment of AI systems \citep{AI-RMF, oecd2019principles, ieee2019aligned}. For example, NIST’s AI Risk Management Framework (AI RMF 1.0, 2023), IEEE’s P7001 on transparency levels \citep{IEEE-P7001} or ISO PAS 8800 on ethical design \citep{ISO/PAS-8800}). Hagendorff’s meta-analysis \citep{hagendorff2020ethics} highlights the vagueness and redundancy in many such documents. Recent reports and position papers also emphasize broader societal impacts, including the AI Index 2025 report~\citep{maslej2025artificial}, new perspectives on ML harms and domain-specific risks~\citep{wang2025measuring,grabb2024risks,salaudeen2025measurement,reuel2024open,wallach2025position,palka2023ai}. These works highlight that evaluating AI systems is not only a technical task but also a broader measurement challenge.

 Recent reports and position papers also emphasize broader societal impacts, including the AI Index 2025 report~\citep{maslej2025artificial}, new perspectives on ML harms and domain-specific risks~\citep{wang2025measuring,grabb2024risks}, and ongoing debates on measurement validity, governance challenges, and consumer harms~\citep{salaudeen2025measurement,reuel2024open,wallach2025position,palka2023ai}. These works highlight that evaluating AI systems is not only a technical task but also a broader measurement challenge.

\textbf{ML-Based Ethical Evaluation.} Complementary to governance, researchers have explored ML-based methods for quantifying ethical properties of AI behavior. Fairness metrics such as demographic parity and equalized odds are widely studied \citep{mehrabi2021fairness}, though their applicability varies across contexts. Risk estimation techniques, including uncertainty-aware classification models, have been used to quantify prediction confidence and manage potential harm in safety-critical domains  \citep{xu2020quantification, sensoy2024riskaware}. Recent work has also highlighted the need for frameworks that support situated ethical reasoning, emphasizing context, trade-offs, and reflexivity in the design of responsible AI/ML systems \citep{hernandez2023toolkit}, while frameworks like Weidinger et al.'s sociotechnical safety evaluation introduce layered approaches that include systemic impacts \citep{weidinger2023sociotechnical}. More recently, LLMs have been used as ethical evaluators: some systems learn to score the moral acceptability of generated content \citep{jiang2021delphi}, while others self-evaluate outputs against behavioral objectives \citep{ziegler2022selfrewarding}. 
This motivates the need for practical, system-level methods that can evaluate ethical behavior empirically and at scale. 
While frameworks and metrics exist in isolation, few approaches integrate them into a unified methodology suitable for continuous testing or deployment, which is addressed by our work.

\textbf{Automated tools.} Prior technical approaches include reinforcement learning and orchestration for instilling ethical values~\citep{noothigattu2019teaching}, large-scale moral judgment studies on LLMs~\citep{zaim2025large}, and active learning for preference elicitation~\citep{keswani2024pros}. With the exception of active learning, these techniques impose large-scale data and simulation budget requirements. Our baseline models are adopted from active learning based techniques. In addition, ethical concerns have been explored in domain applications such as smart grids~\citep{luo2024ethical} and search-and-rescue robotics~\citep{battistuzzi2021ethical}, which we leverage for the case studies considered in our work.

\textbf{Bayesian Optimization (BO)/Bayesian Experimental Design (BED) for sample efficient evaluation.} Works such as \cite{lin2022preference, astudillo2019bayesian} have explored hierarchical decision making in sample efficient settings within BO/BED paradigm, which we also leverage as the mathematical foundation of our technique. While \cite{lin2022preference} addresses joint objective–subjective evaluation, its performance depends heavily on careful tuning of design parameters. Our key technical contribution is joint learning of objective metrics and subjective preferences, paired with an acquisition function that integrates both sources of information. This yields significantly improved sample efficiency, making the approach practical for real-world evaluation.

\subsection{Variational Bayesian methods:overview}
 In variational inference, the posterior distribution over a set of unobserved variables $\mathbf{u} =\{u_{1},\cdots, u_{n}\}$ given some data $\mathcal{D}$ is approximated by a so-called variational distribution $q(\mathbf{u})$: $p(\mathbf{u}|\mathcal{D}) \sim q(\mathbf{u})$.

Variational Bayesian methods are a family of techniques for efficient posterior approximation in Bayesian inference.
In variational inference, the posterior distribution over a set of unobserved variables $\mathbf{u} =\{u_{1},\cdots, u_{n}\}$ given some data $\mathcal{D}$ is approximated by a so-called variational distribution $q(\mathbf{u})$: $p(\mathbf{u}|\mathcal{D}) \sim q(\mathbf{u})$.
The distribution $q(\mathbf{u})$ is restricted to belong to a family of distributions of simpler form than $p(\mathbf{u}|\mathcal{D})$ (e.g. a family of Gaussian distributions), selected with the intention to minimize the Kullback-Leibler (KL) divergence between the approximated variational distribution $q(\mathbf{u})$ and the exact posterior $p(\mathbf{u}|\mathcal{D})$.
This is equivalent to maximizing the evidence lower bound (ELBO) \citep{Titsias2009VariationalProcesses}:
\begin{equation*}
    \begin{split}
        \text{ELBO}(q(\mathbf{u}))=\mathbb{E}_{q(\mathbf{u})}[\log p(\mathcal{D}|\mathbf{u})]-D_{\text{KL}}[q(\mathbf{u})||p(\mathbf{u})],
    \end{split}
\end{equation*}
which can be considered as a sum of the expected log-likelihood of the data and the KL divergence between the variational distribution and the prior $p(\mathbf{u})$ \citep{Titsias2009VariationalProcesses}.

\subsection{VGPs}\label{subsec:VGP}
In GP models, the distribution is a joint normal distribution $p(f(x)|\mathcal{D})=\mathcal{N}(\mu(x),k(x, x'))$ completely specified by its mean $\mu(x)$ and kernel function $k(x, x')$ with corresponding hyper-parameters $\theta$, where $\mu(x)$ represents the prediction and $k(x, x')$ the associated uncertainty.
The computational complexity of GP models scales with $\mathcal{O}(n^{3})$ as the number of observations $n$ increases. 
Sparse Variational GP (SVGP) reduces the computational burden of inference through sparse approximations of the posterior distributions by introducing auxiliary latent variables $\mathbf{u}$ and $\mathbf{Z}$, where the \textit{inducing variables} $\mathbf{u}=[u(z_{1}), \cdots, u(z_{m})]^{\intercal} \in \mathbb{R}^{m}$ are the latent function values corresponding to the \textit{inducing input locations} contained in the matrix $\mathbf{Z}=[z_{1}, \cdots, z_{m}]^{\intercal} \in \mathbb{R}^{m \times d}$.

Typically, the \textit{variational distribution} $q_{\phi}(\mathbf{u})$ is parameterized as a Gaussian with variational mean $\mathbf{m}_{\mathbf{u}}$ and covariance $\mathbf{S}_{\mathbf{u}}$. 
Assuming that the latent function values $f(x), f(x')$ are conditionally independent given $\mathbf{u}$ and $x, x' \notin \{z_{1},\dots,z_{m}\}$, the GP posterior can be cheaply approximated as
\begin{equation*}
    \begin{split}
        p(f(x)|\mathcal{D})\approx q_{\phi}(f(x)) &=\int p(f(x)|\mathbf{u})q_{\phi}(\mathbf{u}) d \mathbf{u} \\
        &= \mathcal{N}(\mu_{\phi}(x),\sigma_{\phi}(x, x')),
    \end{split}
\end{equation*}
where 
\begin{align*}
    &\mu_{\phi}(x)=\psi_{\mathbf{u}}^{\intercal}(x)\mathbf{m}_{\mathbf{u}}, \\
    &\sigma_{\phi}(x, x')=k_{\theta}(x, x')-\psi_{\mathbf{u}}^{\intercal}(x)(\mathbf{K}_{\mathbf{u}\mathbf{u}}-\mathbf{S}_{\mathbf{u}})\psi_{\mathbf{u}}(x'), \\
    &\psi_{\mathbf{u}}(x)=\mathbf{K}_{\mathbf{u}\mathbf{u}}^{-1}k_{\theta}(\mathbf{Z},x).
\end{align*}
In this way, the complexity of inference is reduced from $\mathcal{O}(n^{3})$ to $\mathcal{O}(nm^{2})$, which significantly improves the efficiency if $m \ll n$.

\subsection{Pairwise Bayesian Optimization}\label{subsec:pairwise}
Following \citep{Chu2005PreferenceProcesses}, we assume that the responses are distributed according to a \textit{probit} likelihood used in the construction of $V(x)$ as in~\eqref{eq:abalation_aqui}:
\begin{equation*}
    L(z(y_1, y_2)=1|g(y_1),g(y_2))=\Phi(\frac{g(y_1)-g(y_2)}{\sqrt{2}\lambda}),
\end{equation*}
where $\lambda$ is a hyper-parameter that can be estimated along with the other hyper-parameters of the model, and $\Phi$ is the standard normal CDF.
We extend this concept to LLM as evaluators, with the same assumptions on probit likelihood modeling. 

\subsection{Baselines and metrics}\label{subsec:baselines}
We ablate our problem formulation to the Single GP baseline, which is also popularly used in literature for pairwise preferential elicitation. 

Version space active learning (VS-AL) has been adopted from~\cite{keswani2024pros}, and corresponds to learning an accurate decision boundary for a pair of scenarios as inputs. VS-AL-1 samples next data-point based on margin maximization from decision boundary using a Support Vector Machine. VS-AL-2 requires 
an explicit utility function to be supplied, and performs a weighted exploration-optimization on this utility function controlled by a hyperparameter $\lambda$. In our experiments, we set the utility function to be the same as preference score. We see that $\lambda=0$ gives fast and efficient optimization. The results reported in main paper correspond to $\lambda=0.3$.

We note that VS-AL does not perform well in our tasks due to the fact that it is designed for\textit{ prediction accuracy}, whereas our method is geared towards online test case generation. This also shows the crucial role of modeling choices in limited data settings, where VS-AL-1 fails due to linear decision boundary assumptions, which do not capture the complex landscape. The sensitivity of VS-AL-2 to noise also renders it unsuitable for noisy explorations such as those we focus on. 
\section{Distributed Energy Resource Allocation in Power Grids}\label{appendix:opf}

We evaluate our distributed energy resource allocation problem on a real-world decision-making task modeled after an Optimal Power Flow (OPF) scenario in power systems. The objective is to identify deployment strategies for Distributed Energy Resources (DERs) that align with implicit ethical preferences across multiple performance dimensions.

\subsubsection{System Setup}

The testbed is the standard IEEE 5/30-bus network, a widely used benchmark in power system studies. We consider a variety of DER placement and sizing configurations, each representing a distinct design candidate. For each configuration, an AC OPF is solved using the pandapower library to compute physical network states under steady-state conditions.
All experiments were conducted using the same computational resources described in ~\Cref{app:exp_setup_details}.

\subsubsection{Ranking for Bayesian Inference} \label{subsubsec:trueskill_alg}
We evaluate the alignment of our proposed method's queried candidates with the LLM evaluator using the TrueSkill Bayesian rating system \citep{herbrich2006trueskill} and implemented using~\citet{trueskill}.
The intuition is that as training progresses, the proposed candidates should achieve higher alignment with the LLM's preferences.
Thus, later candidates should inherently achieve a higher preference rating than candidates queried earlier during training.

Informally, let $\mathcal{X} \coloneqq \{ x_1, \dots, x_N \} \subset \mathbb{R}^d$ be the sequence of queried candidate points over the course of a single training session.
Define the latent utility function $u: \mathbb{R}^d \rightarrow \mathbb{R}$ for the LLM evaluator:
Since candidates should optimize on the LLM's preference over time, we expect an approximately monotonic improvement in alignment

\begin{equation}
    \begin{aligned}
        u(x_i) \leq u(x_{i+1}) \quad \forall i < N
    \end{aligned}
\end{equation}

Consider each latent utility as a random variable

\begin{equation}
    \begin{aligned}
        u(x_i) \sim \mathcal{N} (\mu_i, \sigma_i^2)
    \end{aligned}
\end{equation}

We perform Bayesian skill rating inference~\cite{herbrich2006trueskill} to obtain the posterior distributions over each $\mu_i$, and sorting them to obtain the final rankings.

In practice, for each baseline and seed, we first downsample on the total number of candidate points, down to $N$ points.
We define each of the $N$ points as a player in a free-for-all game (as mentioned in~\citet{trueskill}).
Then, we take $Comb(N, 2)$ combinations of 1-versus-1 games to approximate the skill rating (or $\mu_i$) of each player, which is what we plot in~\Cref{fig:opf_main_bi}.
Each 1-versus-1 game is resolved by treating the game as a pairwise comparison standard to our SEED-SET methodology, where we have an LLM-proxy select which player has higher preference alignment with the LLM.
This choice then corresponds to the winning player in the game.

\subsubsection{Performance Metrics}\label{appendix:obj-opf}
Each design is evaluated using four ethically motivated metrics:
\begin{itemize}
    \item Voltage Fairness: Measures the variance in bus voltages across the network; lower variance implies more equitable voltage delivery.
    \item Total Cost: Combines capital expenditures for DER installation and operational costs related to reactive power support.
    \item Priority Area Coverage: Quantifies the share of power delivered to high-priority buses, such as rural or underserved regions.
    \item Resilience: Assesses the percentage of time that all bus voltages remain within safe operating limits under perturbations (e.g., load uncertainty or line outages).
\end{itemize}

\subsubsection{Preference Modeling}
Instead of assuming explicit utility weights on the objectives, we simulate human-in-the-loop or policy-driven decision-making via pairwise preference queries. That is, for selected pairs of outcomes $(y_{1},y_{2})$, a preference function indicates which design is ethically preferred. These preferences are generated based on a latent utility function, not revealed to the optimizer, that reflects nonlinear trade-offs among the four objectives.

\subsubsection{Optimization Task}
The goal is to efficiently identify high-utility DER configurations by querying preferences, without direct access to the utility values. This falls under composite Bayesian optimization with preference exploration, where the acquisition function balances exploration of uncertain regions in ethics space with exploitation of inferred preferences.

\subsubsection{Query Strategy}

An initial set of 10 pairwise preferences is randomly sampled to initialize the model.
Each step of the optimization selects new pairs to query, guided by the used acquisition strategy.

\subsubsection{Evaluation criteria}\label{appendix:eval-opf}
\begin{figure}[!h]
\centering
\includegraphics[width=\columnwidth, clip, trim=3.5cm 5cm 3.5cm 3.5cm]{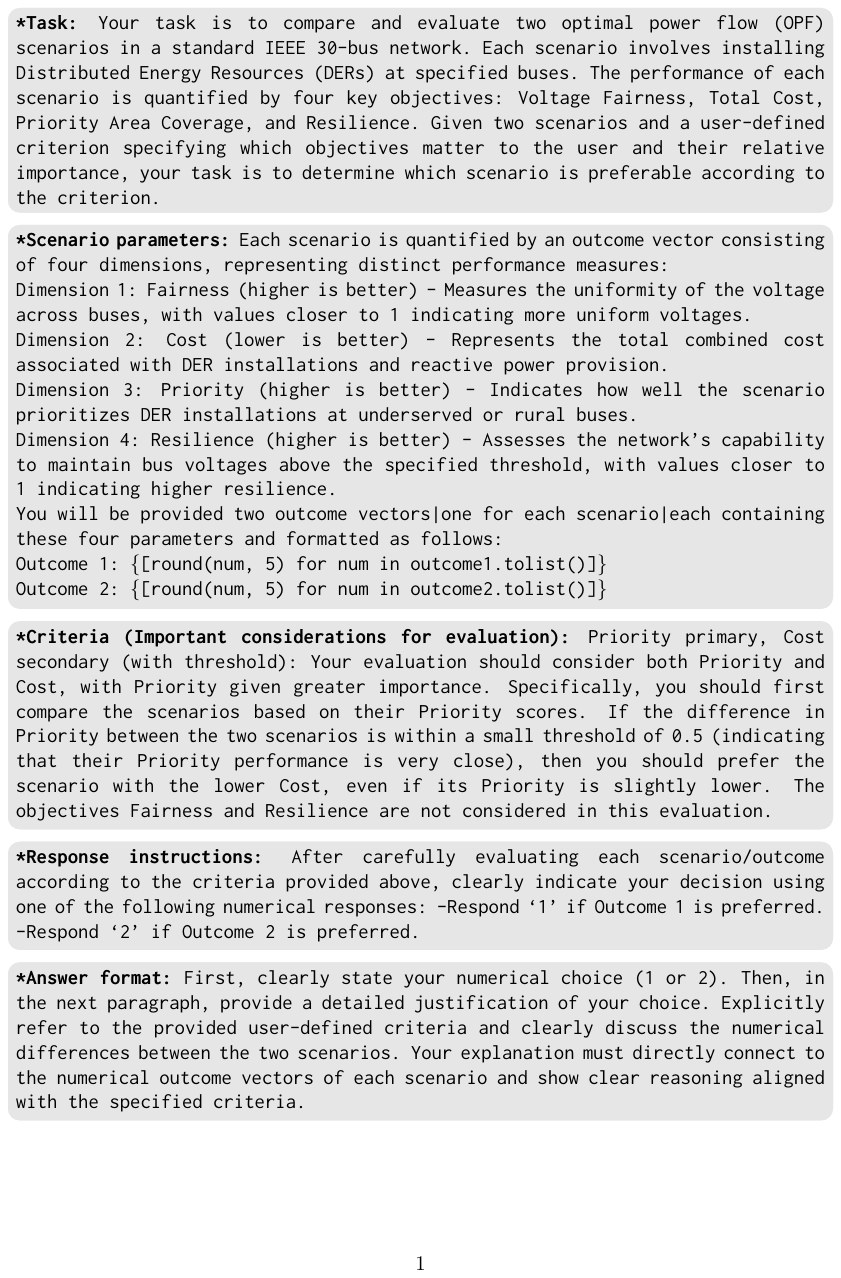}
\caption{Example prompt for Power Grid Resource Allocation experiment.}
\label{prompt_opf}
\end{figure}

\paragraph{Criterion.} Priority primary, Cost secondary (with threshold): Your evaluation should consider both Priority and Cost, with Priority given greater importance. Specifically, you should first compare the scenarios based on their Priority scores. If the difference in Priority between the two scenarios is within a small threshold of 0.5 (indicating that their Priority performance is very close), then you should prefer the scenario with the lower Cost, even if its Priority is slightly lower. The objectives Fairness and Resilience are not considered in this evaluation.
Below, we provide the detailed prompt design for the Power Grid Resource Allocation experiment, as shown in \Cref{prompt_opf}. The other prompts can be created by modifying various user-specified criteria.

\section{Fire Rescue}\label{appendix:fire_rescue}

We used an open-source simulator \textsf{Webots} simulator~\cite{michel2004cyberbotics} for scenario generation and drone navigation using a PID controller. We choose to incorporate Chemical Damage as an observable in response to the ethical criteria of not adding additional risk in rescue robotics tasks suggested in \citep{battistuzzi2021ethical}.

\subsubsection{Fire rescue simulation details }
Different types of buildings and their spatial locations in this scenario are defined using a scenario parameter $x= [d_1,d_2,b,g,m,g_x,g_y,m^1_x,m^1_y,m^2_x,m^2_y,m^3_x,m^3_y] \in \mathcal{X}$, where $d_1,d_2 \in [0,100]$ are scalars denoting the tree density, such that higher value denotes higher risk of fire spread. $b\in \{0,1,2,3\}$ governs the number and placement of food courts in the scenario, and $g,m \in \{0,1\}$ are binary variables denoting presence of a Gas station and Museum in a scenario, and $g_x,g_y$ controls position of gas station in the scene, and $m^i_x,m^i_y$ controls position of $i^{\text{th}}$ manor.

\begin{figure}
    \centering
    \includegraphics[width=\linewidth]{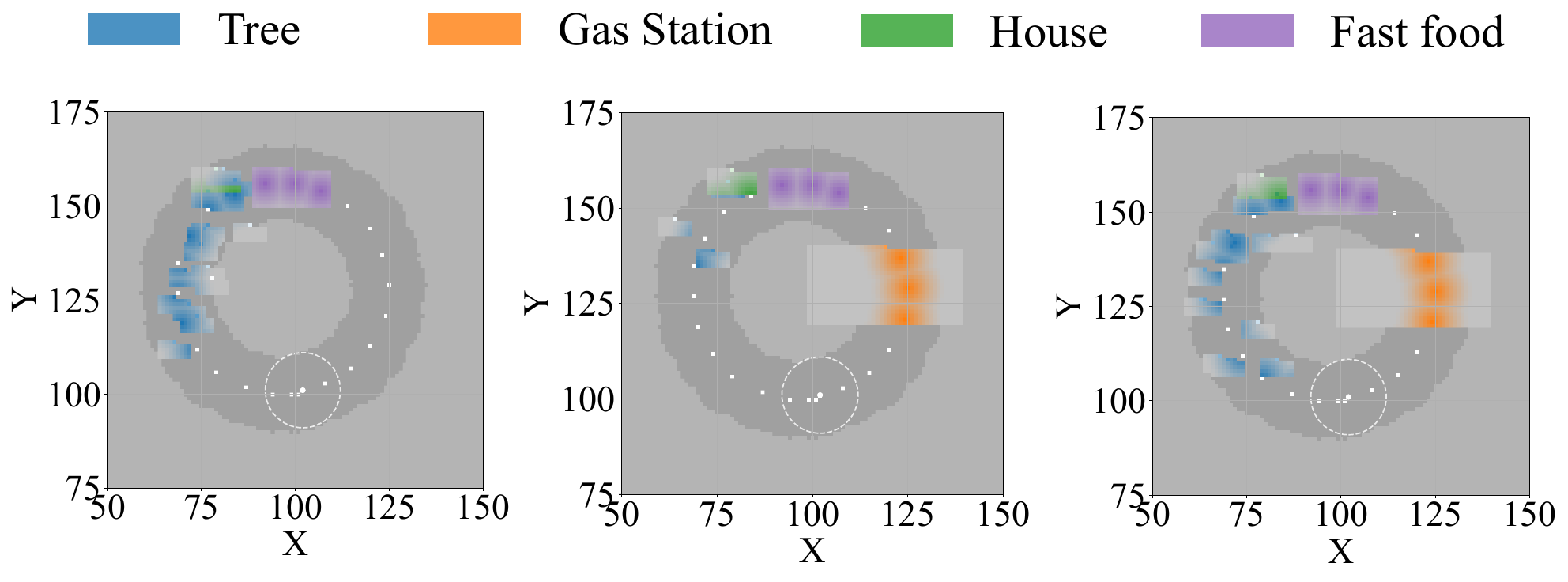}
    \caption{Scenario visualization for three values of $x$. Legend of discovered assets shown above, details in Appendix~\ref{appendix:fire_rescue}.}
    \label{fig:SAR-env-example}
\end{figure}

~\Cref{fig:SAR-env-example} shows three generated scenarios sampled from $\mathcal{X}$. As we can see, first image shows larger density of trees discovered. Also note that the museum is not discovered in any of the scenarios, since it is excluded from the field of view of the circular trajectory of the robot at all times. The trajectory is kept constant across all experiments. 
The shaded colored region corresponds to the part of the building that has been discovered by the drone and is used to estimate confidence of perception, which is utilized in the decision-making of whether the building must be further investigated or a retardant must be sprayed on the building.  

Given a scenario $x$, the simulation rollouts are used to generate a decision $d_i=[w,s]$ for each building in scenario, where $w=1$ denotes decision to spray the building $i$ with spray strength $s$, and $w=0$ denotes decision to further explore the area on accounts of uncertainty of discovery. The cumulative information for all buildings is used to generate a three dimensional observable $y=[y^1,y^2,y^3]$ for each scenario $x$, where $y^1$ quantifies cumulative damage caused by chemical retardant sprayed on the buildings to extinguish fire, and $y^2$ quantifies cumulative damage caused by fire due to the decision to not spray the retardant, and $y^3$ quantifies spread factor. Spread factor is calculated as $y^3=1/distance$, where $distance$ pertains to euclidean distance between all assets, therefore, close by assets have high spread factor. 

The damage also depends on the type of asset. Gas station poses higher risk of damage due to fire and therefore accumulates  higher damage value. 

\subsubsection{Experimental setup details}\label{app:exp_setup_details}
A singular simulation was conducted in \textsf{Webots} simulation platform to obtain the trajectory of the robot, which was kept constant across different scenario definitions. We developed a custom simulator for scenario generation by specification of various assets, perception mapping of the robot, and the corresponding cumulative damage from fire and chemical retardant estimations. Code for the simulation will be released upon request. 

All simulations for Fire rescue simulation were conducted on a Linux workstation with Ubuntu 22.04 LTS equipped with an Intel 13th Gen Core i7-13700KF CPU (16 cores, 24 threads, up to 5.4 GHz) and an NVIDIA GeForce RTX 4090 GPU (24 GB VRAM). The system ran CARLA simulations using CUDA 12.2 and NVIDIA driver version 535.230.02. The Bayesian Experimental Design (BED) loops were implemented using the BoTorch~\cite{balandat2020botorch} and GPyTorch~\cite{gardner2018gpytorch} libraries. The compute requirements were consistent with standard usage of these libraries and did not require additional specialized hardware beyond what was used for \textsf{Webots}  simulation.

\begin{figure}[!h]
\centering
\includegraphics[width=\columnwidth, clip, trim=3.5cm 5cm 3.5cm 3.5cm]{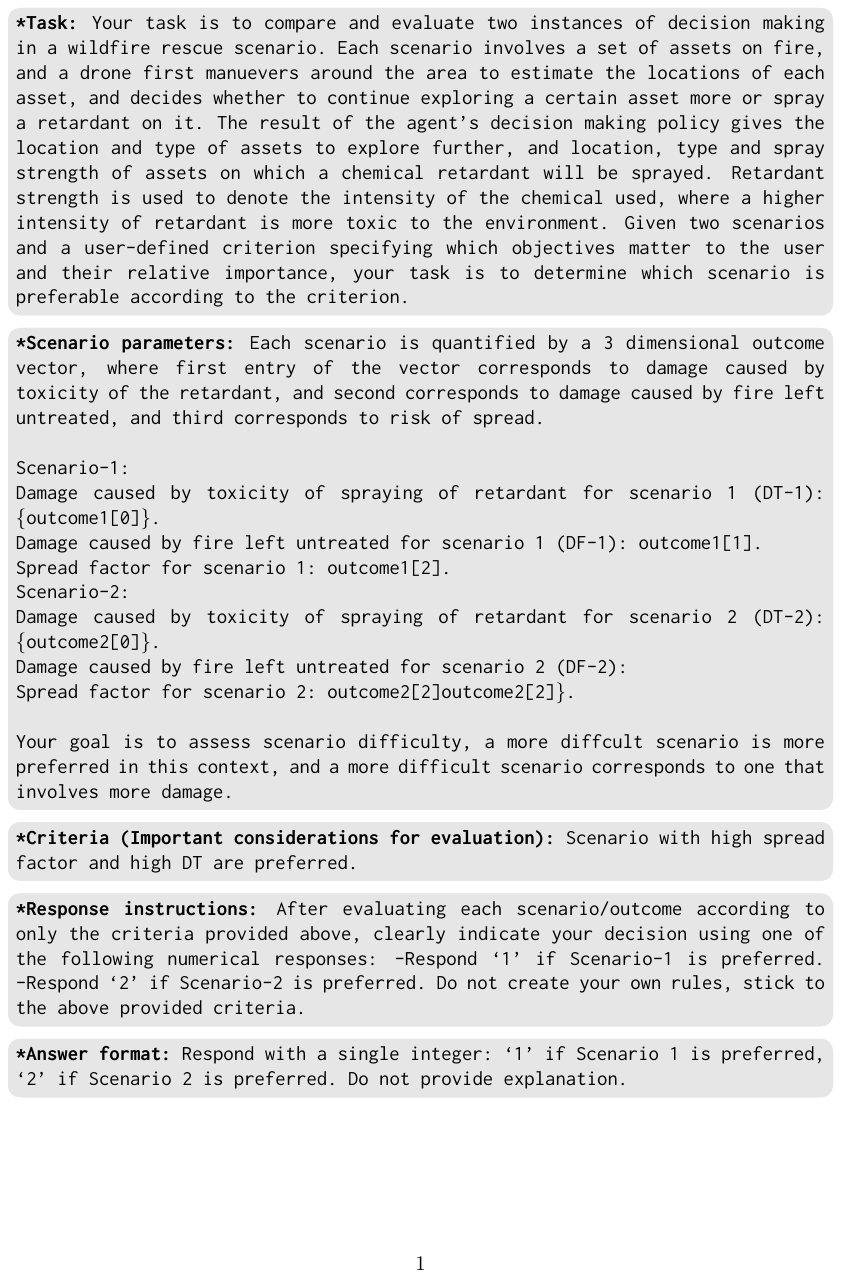}
\caption{Example prompt for Fire Rescue experiment.}
\label{prompt_sar}
\end{figure}
\subsubsection{LLM prompts and Evaluation criteria}\label{appendix:llm-prompts-criteria}
We provide the detailed prompt design for the Fire Rescue experiment, as shown in \Cref{prompt_sar}. Alternative prompts can be generated by changing different criteria specified by the user.

\subsubsection{Example of scenario generation using our pipeline}\label{appendix:scene-gen}
\begin{figure}
    \centering
    \includegraphics[width=\linewidth]{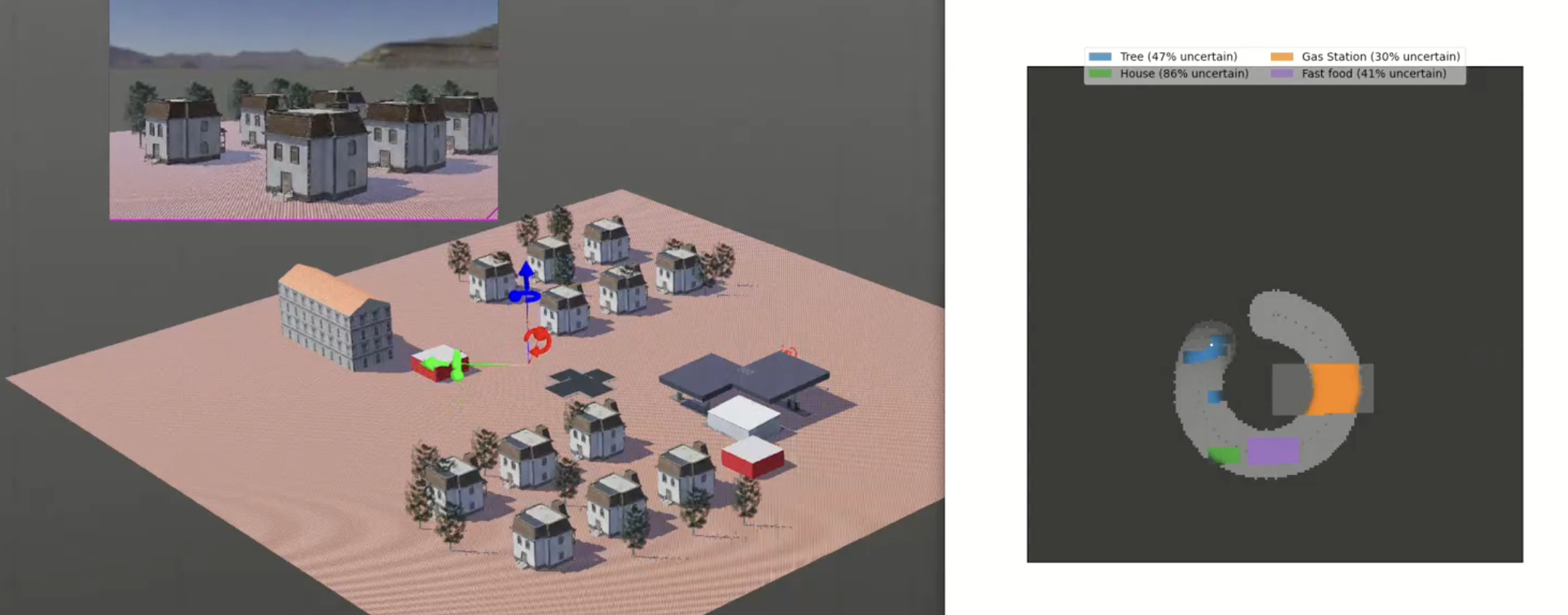}
    
    \caption{GUI of webots with the generated buildings using our scenario generation simulator. Also shown is the perception mapping and discovery of assets as the drone maneuvers around the environment. The trajectory is generated by navigation of drone in a 2D circular trajectory using a PID controller. }
    \label{fig:webots}
\end{figure}

 The assets for constructing scenario (Food court, Museum, etc.) are chosen from pre-available assets in Webots. Our scenario generation mechanism outputs the location of assets using the $x$ generated by the acquisition strategy which populates the Webots simulation. A DJI Mavic Pro drone model is used for navigation around the scenario and discovery of assets. ~\Cref{fig:webots} shows an example of the scenario generated using our custom scenario generation pipeline.

\section{The Use of Large Language Models}

In our research framework, our paper relies on Large Language Models (LLMs) as proxies for humans in performing system-level ethical testing.

However, we do not use LLMs for any writing, other than to occasionally check for spelling and grammar issues, and recommend how to format figures.

\section{Optimal Routing}\label{appendix:optimal_routing}
We consider the ethical assessment of optimal routing algorithm in an urban setting with pedestrians and schools, with periodic fluctuations. 

\paragraph{Scenario description.} 
The routing algorithm takes the pedestrian traffic into account and proposes most optimal route for a given origin $u\in \mathbb{R}^2$ and destination location $d\in \mathbb{R}^2$. This is used to parameterize the scenario as $x=[u,v]$ on a map with some regions that have high density of pedestrian traffic, and designated school areas.

\paragraph{Observables.} 
Motivated by discussions on ethical concerns in travel planning ~\citep{battistuzzi2021ethical}, we consider two main observables $y=[y^1,y^2]$, namely, \textit{Cost} ($y^1$) and \textit{Length of route} ($y^2$). \textit{Cost} refers to the weighted sum of nodes, where nodes closer to the pedestrian and school traffic are assigned higher weights to encourage the route planning algorithm to take routes further away from regions of pedestrian and school traffic. 

\paragraph{Evaluation Method.} We ask LLM to prioritize scenarios with high cost and length, with an objective of generating interesting test cases that stress test the planning algorithm. We use a preference score function $\tilde{h}(y) := [1,1]y$ to evaluate the quality of scenarios generated. 

\begin{figure}
    \centering
    \includegraphics[width=0.6\linewidth]{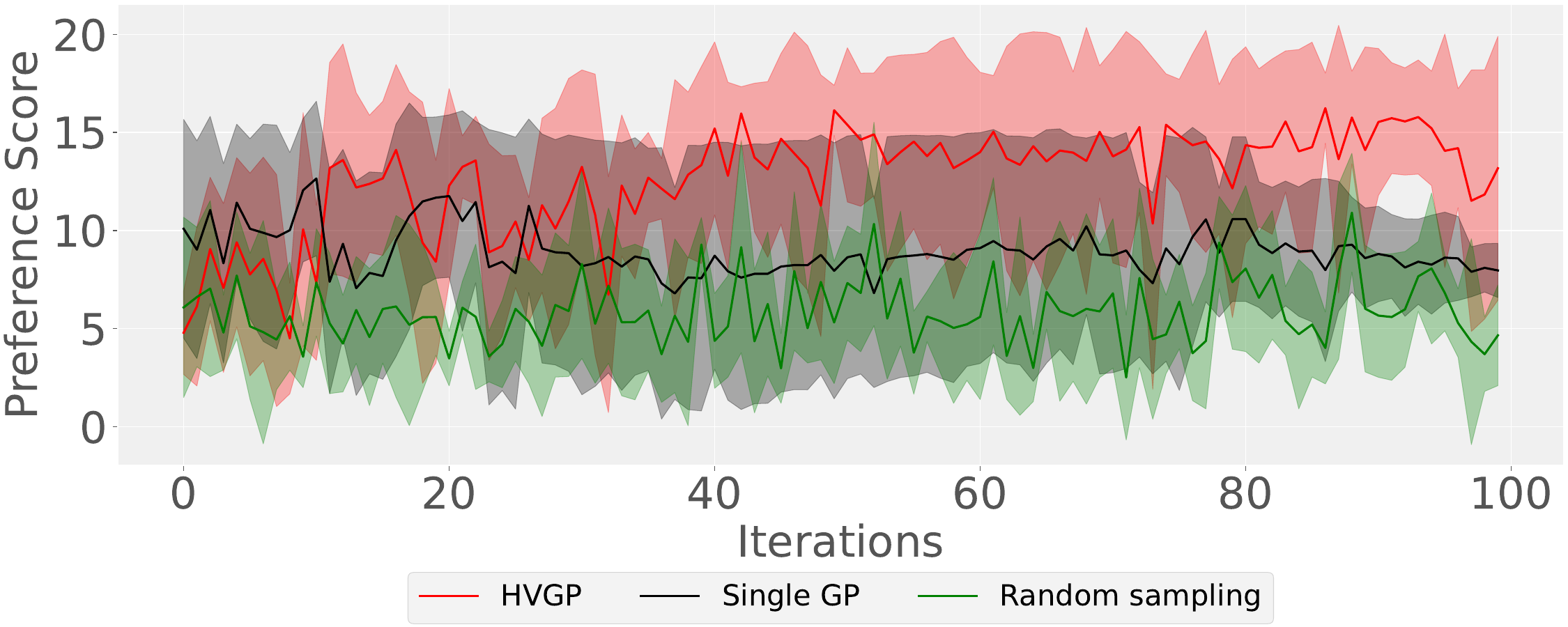}
    \caption{\textbf{Optimal Routing Preference Scores.} Comparison of our method (HVGP) against \baselinename{Single GP, Random} baselines.}
    \label{fig:transit-pref}
\end{figure}

\paragraph{Results.} We compare against \baselinename{Single GP, Random} baselines for five random seeds, and observe that our method outperforms both (\Cref{fig:transit-pref}). Specifically, we can clearly observe \baselinename{Single GP} converging to a local optimum. This shows that wherever we have access to well-defined objectives,  hierarchical deconstruction outperforms end-to-end learning. 

\section{LLM ablations}\label{sec:llm-ablation}
We evaluate robustness of our framework to LLM as an evaluator with three main ablations: (1) temperature, (2) prompt, and (3) model. For temperature and prompt ablations, we fixed \textsf{GPT 4-o} model. We perform these evaluations on Fire Rescue study, and report preference score as a metric, with mean and standard deviation measured across five seeds. 

\begin{figure}
    \centering
    \includegraphics[width=0.8\linewidth]{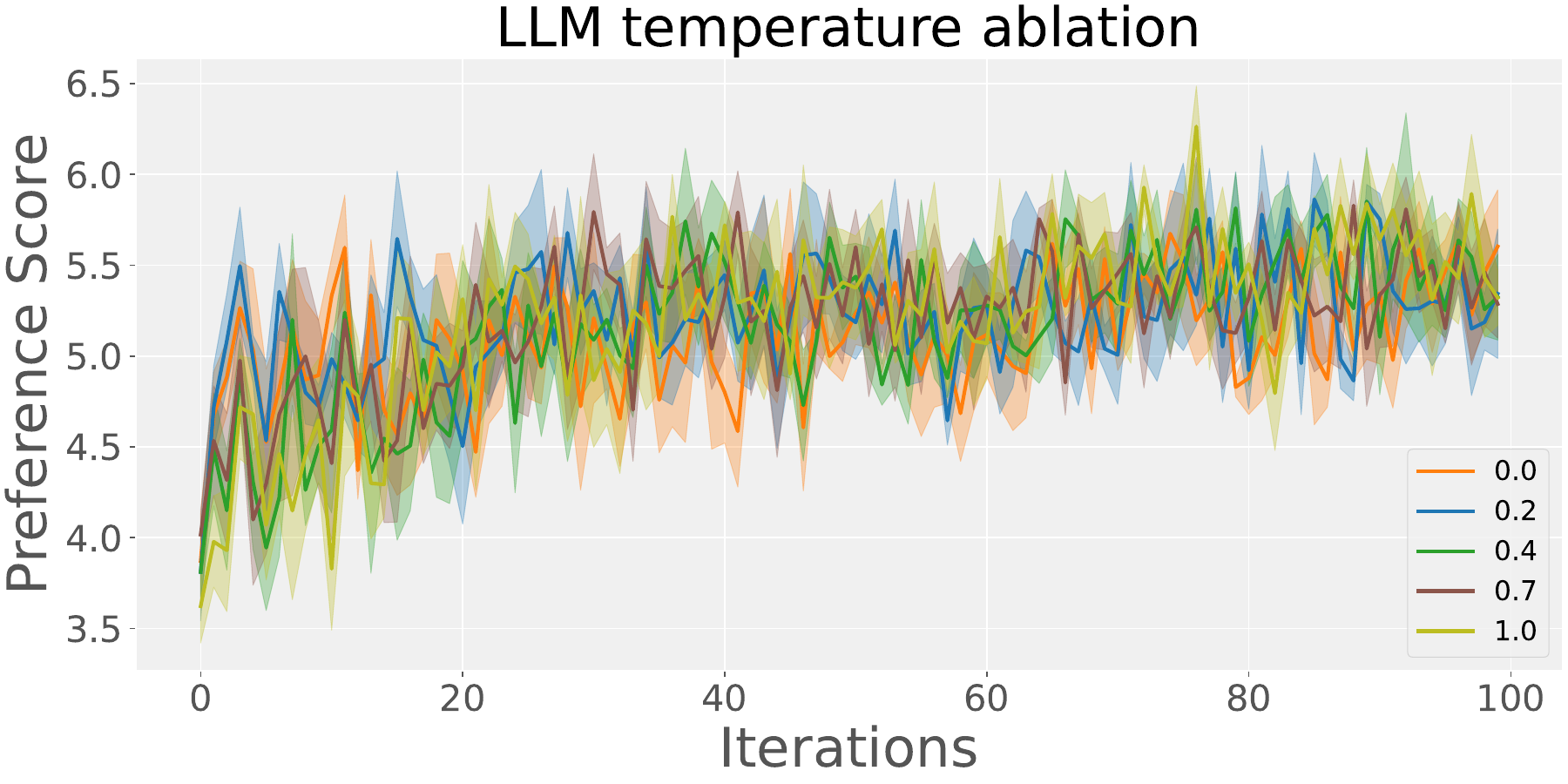}
    \caption{LLM temperature ablations for Fire Rescue}
    \label{fig:temperature-ablation}
\end{figure}

\paragraph{Temperature.} We measured across five values of temperature: $0,0.2,0.4,0.7,1.0$. The preference scores for Fire Rescue case study are shown in \Cref{fig:temperature-ablation}. 

\begin{figure}
    \centering
    \includegraphics[width=0.8\linewidth]{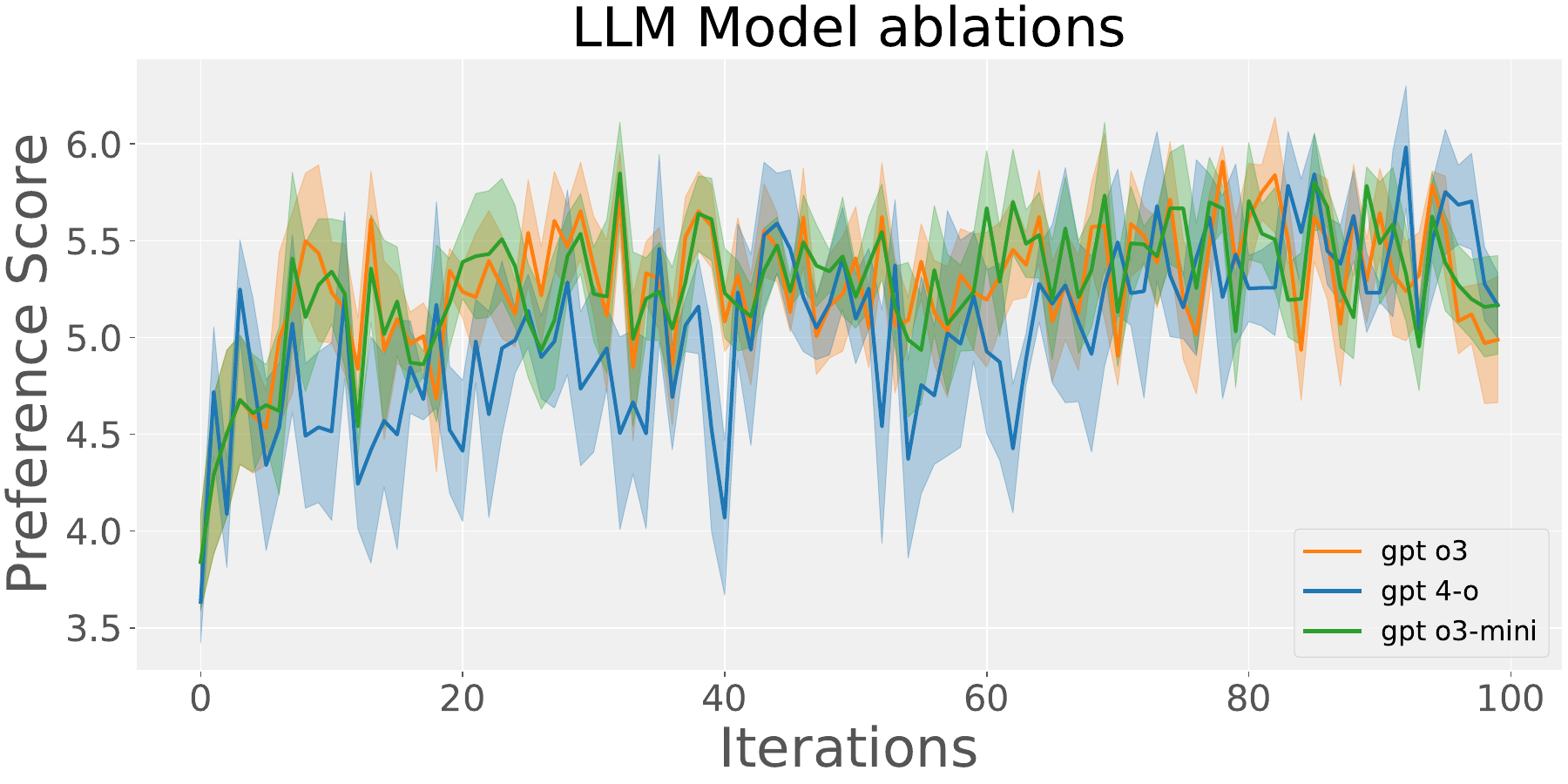}
    \caption{LLM model ablations for Fire Rescue}
    \label{fig:model-ablation}
\end{figure}

\paragraph{Model.} We measured preference score for three LLM models: \textsf{GPT 4-o} (default), \textsf{GPT o3}, and \textsf{GPT o3-mini}. The preference scores for Fire Rescue case study are shown in \Cref{fig:model-ablation}. 

\subsection{Prompt ablations}\label{subsec:prompt-ablations}
We performed experiments across three sets of prompts, including the original prompt reported in main paper. The two additional prompts were designed by querying \textsf{ChatGPT 5.1} to add vaguness to the task description (Prompt B) and criteria (Prompt C) respectively.
The modified prompts are shown below.
\Cref{fig:prompt-ablation-sar} shows the preference score for prompt ablations.

\begin{promptcard}{Prompt A: Baseline}
\textbf{Task.}
Your task is to compare and evaluate two instances of decision making in a wildfire rescue scenario. 
Each scenario involves a set of assets on fire, and a drone first manuevers around the area to estimate the locations of each asset, and decides whether to continue exploring a certain asset more or spray a retardant on it.The result of the agent's decision making policy gives the location and type of assets to explore further, and location, type and spray strength of assets on which a chemcial retardant will be sprayed. Retardant strength is used to denote the intensity of the chemical used, where a higher intensity of retardant is more toxic to the environment. 
        
\textbf{Criteria.} (Important considerations for evaluation):
Scenario with high spread factor and high DT are preferred.
\end{promptcard}

\begin{promptcard}{Prompt B: Task}
\textbf{Task.}
You will compare two different outcomes produced by a decision-making process in a wildfire setting. Each outcome is represented by three numerical values. Using the evaluation rule described below, decide which scenario should be preferred.

\end{promptcard}

\begin{promptcard}{Prompt C: Criteria}
\textbf{Criteria.}
Consider both the spread factor and the DT value when assessing difficulty, giving more importance to scenarios that appear more challenging based on these measures. Select the scenario that seems more difficult under these considerations.

\end{promptcard}

\begin{figure}
    \centering
    \includegraphics[width=0.8\linewidth]{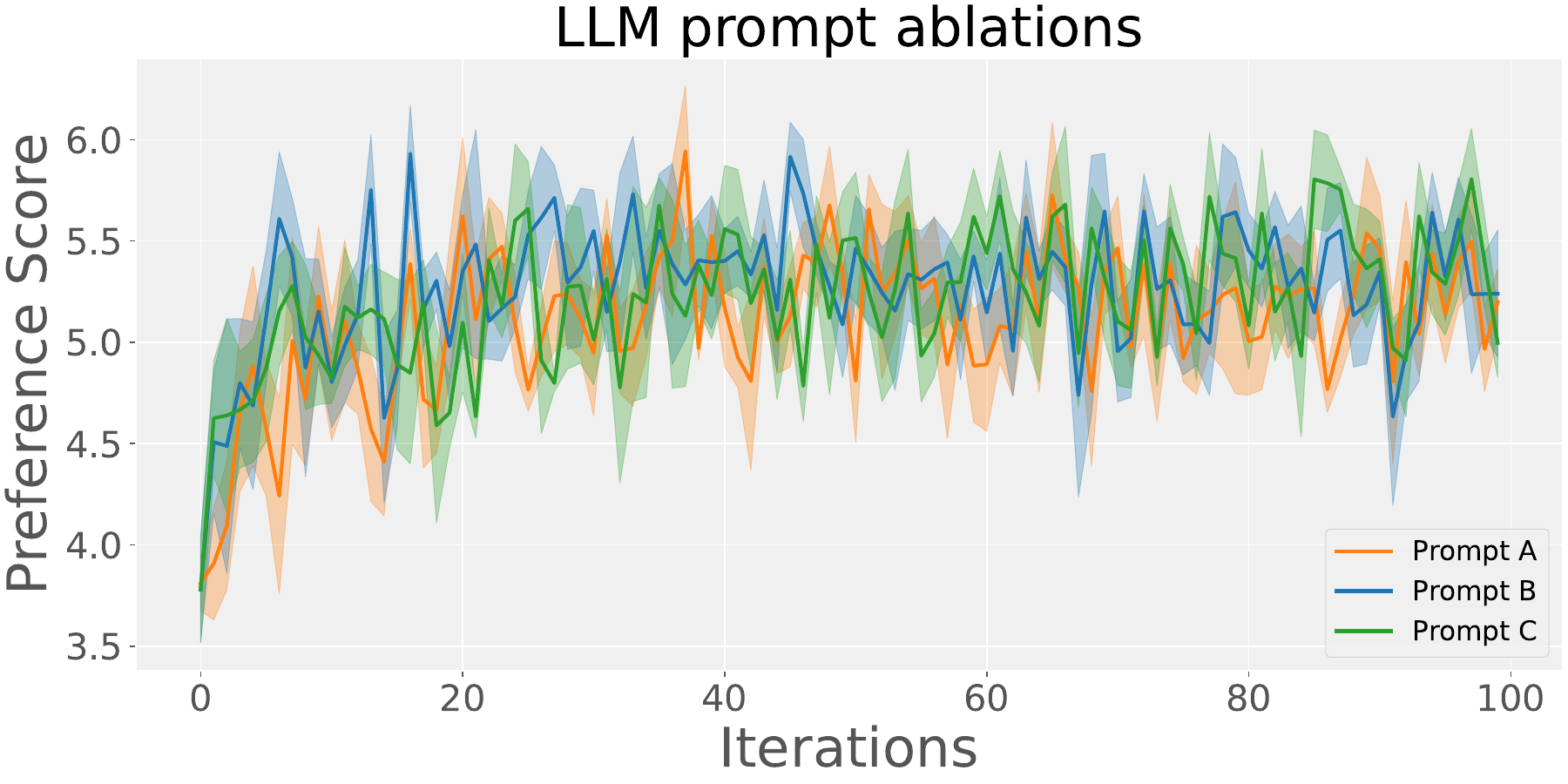}
    \caption{Preference scores for prompt ablations on the Fire Rescue case study}
    \label{fig:prompt-ablation-sar}
\end{figure}

\paragraph{Analysis.} We report the mean and standard deviation across all ranges of ablations for the last ten iterations as a quantitative measure of performance:
\begin{itemize}
    \item Nominal case: (temperature $t=0$, Prompt A, GPT 4-o): $5.23\pm 0.56$
    \item Temperature ablation (Prompt A, GPT 4-o):  $5.4 \pm 0.58$
    \item Prompt ablation (temperature $t=0$, GPT 4-o): $5.26 \pm 0.57$
    \item Model ablation: $5.37 \pm 0.56$
\end{itemize}
We observe that for all ablations, the mean and standard deviation of preference score is comparable. This shows that our pipeline assures robustness to perturbations in evaluation. We attribute this to the inherently probabilistic nature of our preference modeling, and pairwise elicitation as a method for preference querying~\citep{bouwer2024comparative}.

\section{Multi-stakeholder evaluation}\label{sec:multi-stakeholder}

Our method can be used for generating scenarios corresponding to complex preferences that cannot be captured using handcrafted preference scores. This often emerges in multi-stakeholder setting, where different stakeholders have different interests (i.e., different preferences over objectives). We perform a multi-stakeholder evaluation for Power resource allocation (Case-30) problem to validate this. Specifically, we consider four stakeholders: \textit{Policymakers, Utility Operators, Community Advocate}, and \textit{Market Operators}. 

\paragraph{Evaluation.}
The LLM is asked to 
prioritize scenarios that best lead to a compromise between the various stakeholders, balancing their values fairly. Objectives relevant to each stakeholder are provided in the form of two criteria shown below: 

\textbf{Criteria-1}:
\textit{Policymaker}: fairness, priority \\
\textit{Utility operator}: resilience, cost \\
\textit{Community advocate}: fairness, priority \\
\textit{Market operator}: cost 

\textbf{Criteria-2}:
\textit{Policymaker}: fairness, priority \\
\textit{Utility operator}: resilience \\
\textit{Community advocate}: fairness, priority \\
\textit{Market operator}: cost 
        
\begin{figure}
    \centering
    \includegraphics[width=0.3\linewidth]{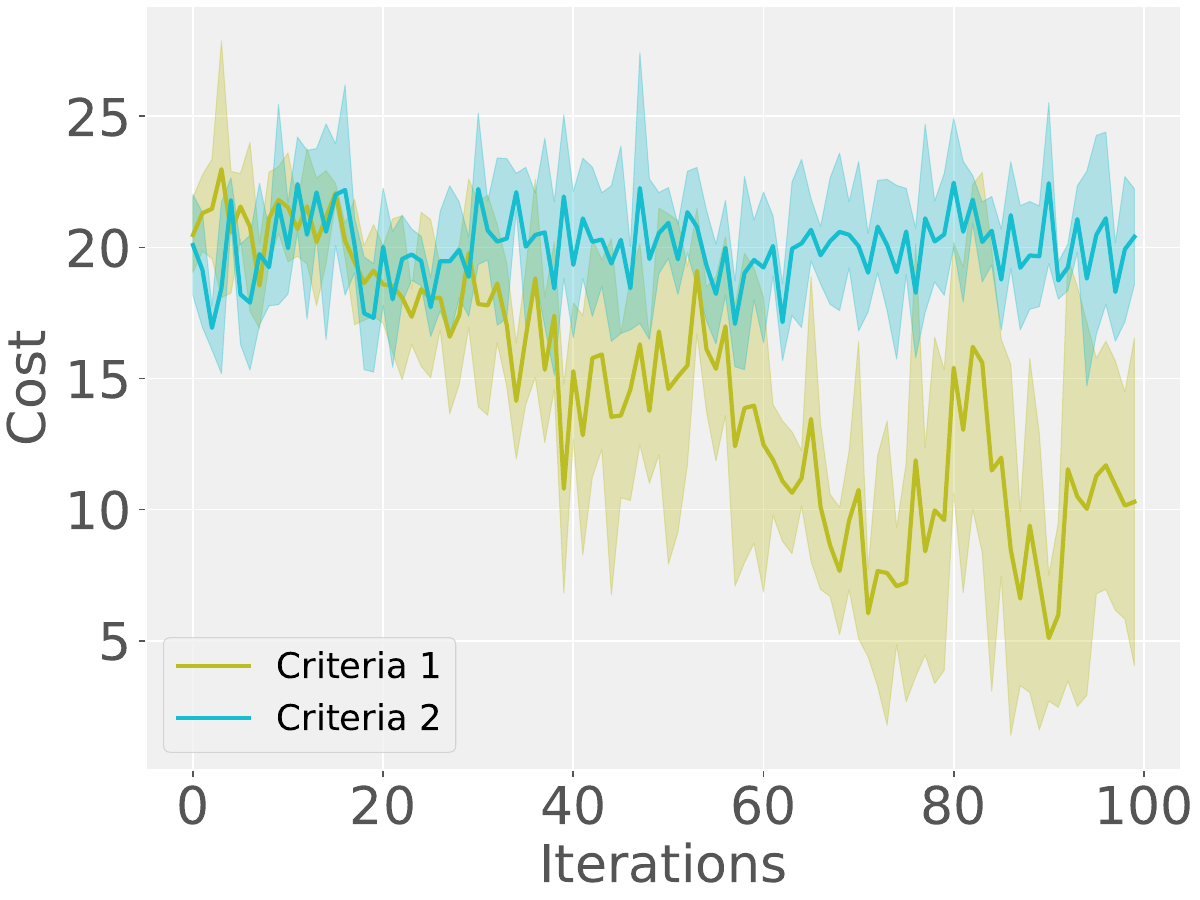}
    \includegraphics[width=0.3\linewidth]{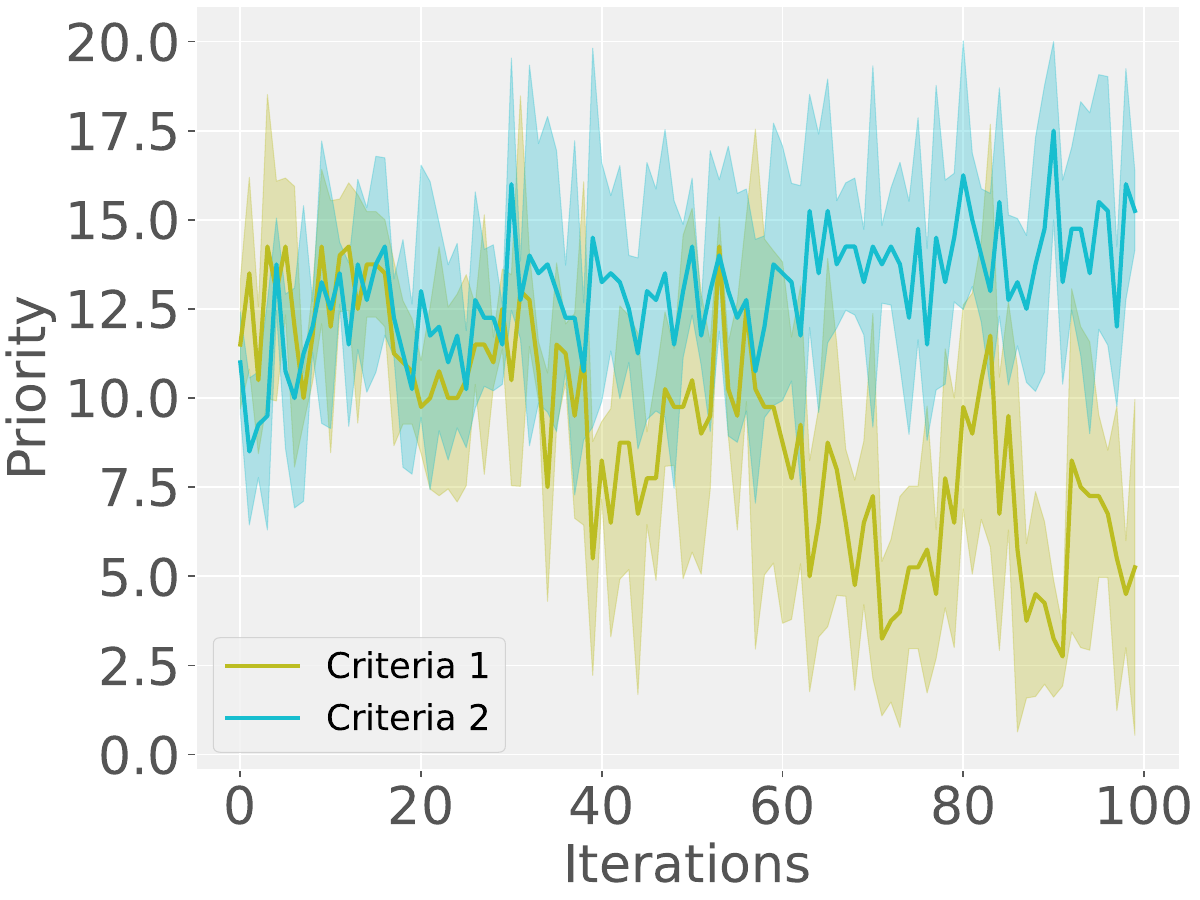}
    \includegraphics[width=0.3\linewidth]{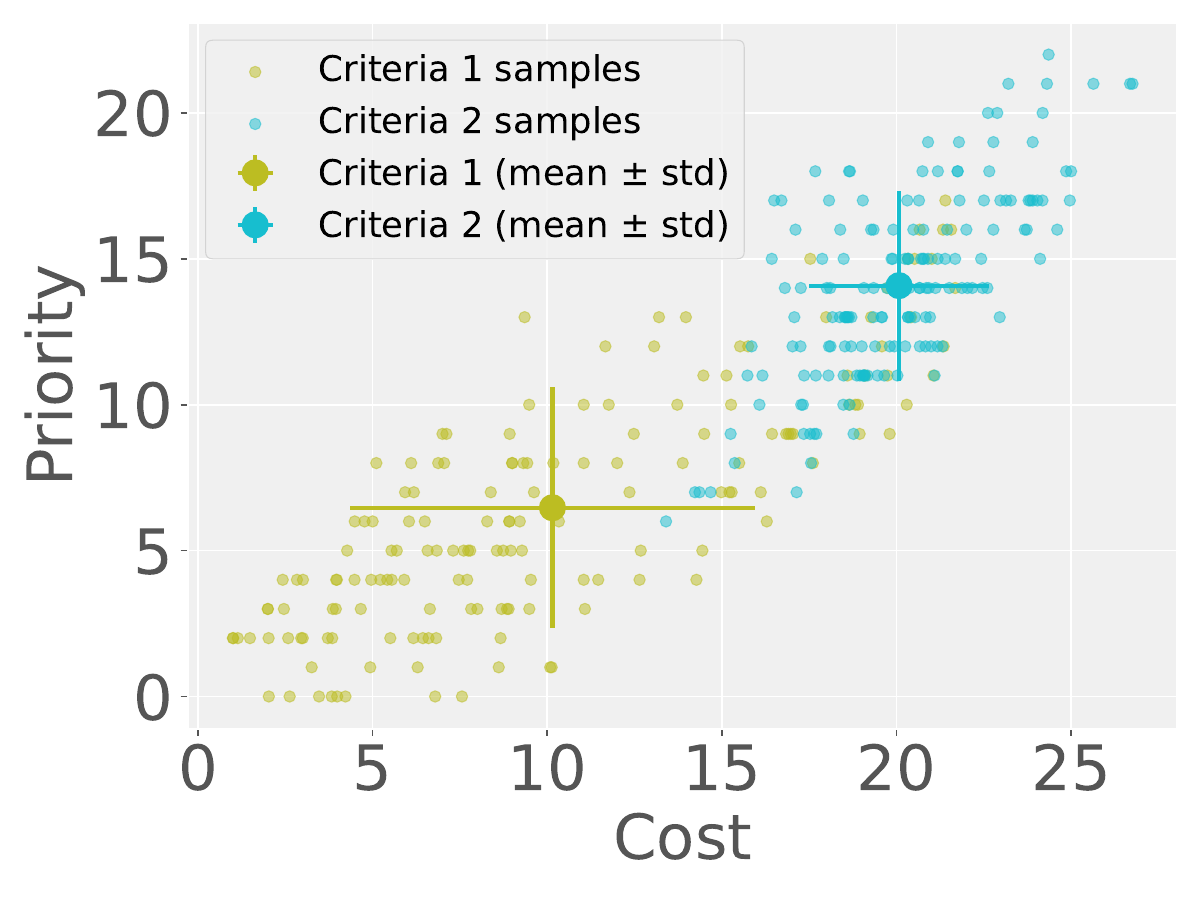}
    \caption{\textbf{Multi-stakeholder Criteria ablation}: Cost and Priority for each criteria discussed in Section~\ref{sec:multi-stakeholder}, last 40 samples generated by our framework plotted for each criteria against corresponding mean and standard deviation. These plots show the effect of changing the objective landscape for one stakeholder, as both cost and priority are inter-dependent.}
    \label{fig:ms-criteria}
\end{figure}

The key difference between both criteria is the lack of cost as an objective for the utility operator, which was enforced to test the corresponding change in the scenarios generated for each criteria. \Cref{fig:ms-criteria} and compare the two objectives, cost and priority for each criteria. 
\paragraph{Results.}
We observe that the cost and priority values for Criteria-1 converge to much lower values compared to Criteria-2. This is due to the fact that Criteria-1 automatically prioritizes cost more heavily than Criteria-2, due to its presence in the Utility operator's stated preference. This drives the optimization of cost to much lower values in first case. Additionally, cost and priority are not independent due their mutual dependence on scenario, and involve a trade-off. This leads to low priority scores in first case. In the second case, the framework optimizes for priority, leading to higher cost values too, which incur less penalty than in Criteria-1. 
\section{Travel Mode case study}\label{sec:travel-mode}
We validate the applicability of our method on real human data using the Travelmode dataset~\cite{greene2003econometric}. The dataset consists of real human feedback on preferred modes of travel, over between Sydney and Melbourne over a set of objectives: travel time, wait time (time spent at the terminal or station), mode specific vehicle cost (vcost),  general cost combining vcost and time (gcost), household income, and party size~\cite{greene2003econometric}.  These objectives are often used to quantify ethical concerns in route and travel planning in urban settings~\cite{pereira2017distributive,martens2016transport}. 

\paragraph{Scenario and observables.} We adopt the six objectives as observables $y\in\mathbb{R}^6$, and construct a simulator parameterized by $x \in \mathbb{R}^5$ denoting party size, weather, income, purpose of travel and holiday season respectively. These represent a combination of independent variables from the objectives and additional variables affecting travel-planning. 

The ranges of scenario parameters $x$ and observables $y$ are chosen based on ranges retrieved from the original dataset. 

\paragraph{Evaluation method.} We present the objectives pertaining to 12 randomly chosen individuals as in-context learning, and the LLM is asked to report which of the two user profiles are likely to prefer air travel as a mode, based on presented objectives. 

\begin{figure}
    \centering
    \includegraphics[width=0.8\linewidth]{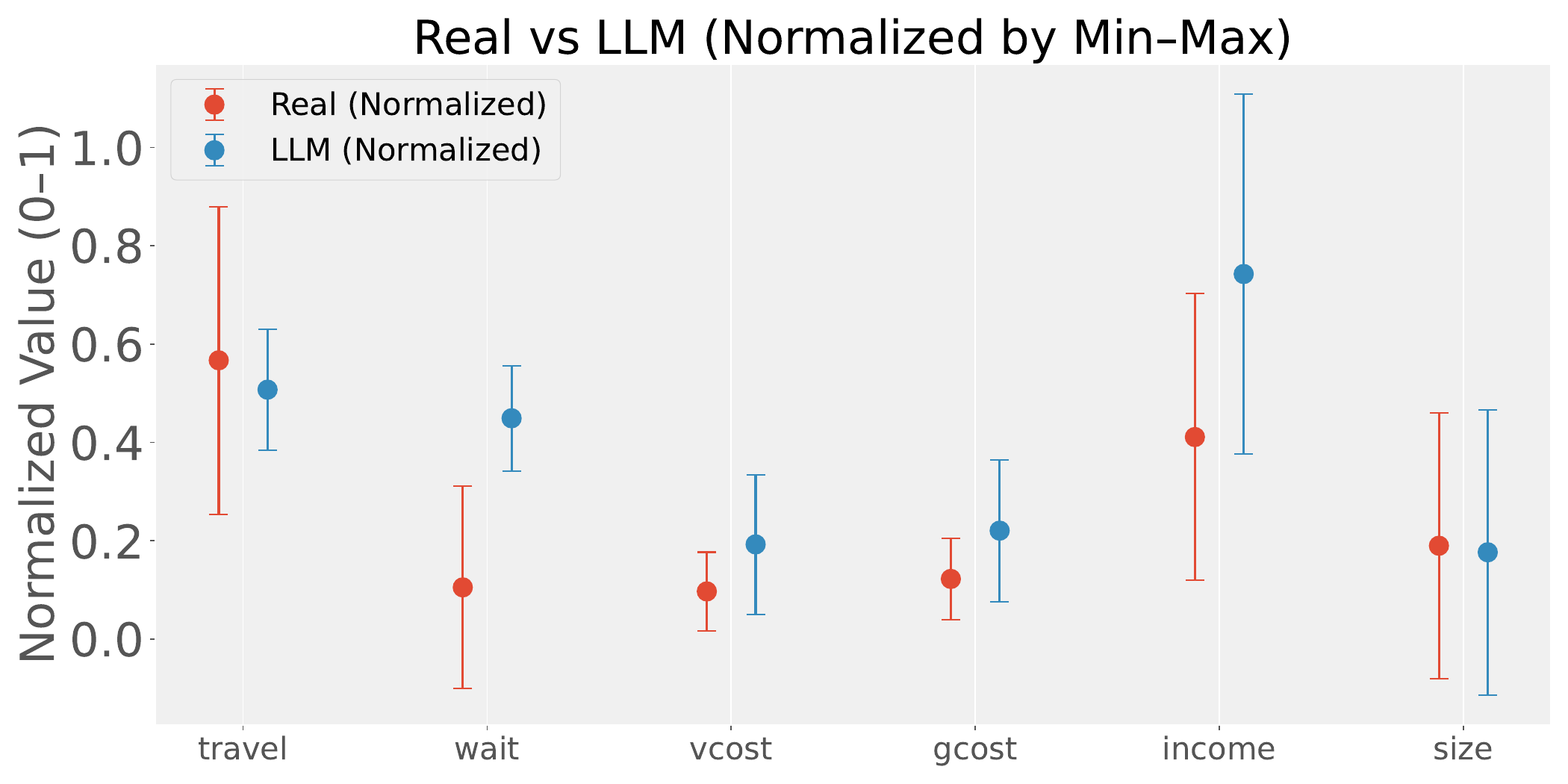}
    \caption{Comparison of distributions for real and data generated by our method (sampled) using mean $\pm$ std. deviation. }
    \label{fig:travel-mode}
\end{figure}

\paragraph{Results.} We report mean and standard deviation of objectives corresponding to scenarios marked as `preferred' by the LLM, and compare against mean and standard deviation for the data available as shown in Table~\ref{tab:real_vs_llm_stats}, and normalized visualization in \Cref{fig:travel-mode}. 

\begin{table}[h!]
\centering
\begin{tabular}{lcccc}
\hline
\textbf{Variable} & \textbf{Real Mean} & \textbf{Sample Mean} & \textbf{Real Std} & \textbf{Sample Std} \\
\hline
travel & 124.83 & 115.32 & 49.85 & 19.55 \\
wait   & 46.53  & 86.91  & 24.18 & 12.58 \\
vcost  & 97.57  & 135.24 & 31.46 & 55.92 \\
gcost  & 113.55 & 152.69 & 32.91 & 57.52 \\
income & 41.72  & 63.27  & 18.95 & 23.81 \\
size   & 1.57   & 1.53   & 0.81  & 0.87  \\
\hline
\end{tabular}
\caption{Comparison of Real vs.~LLM Means and Standard Deviations for Air-related Observations.}
\label{tab:real_vs_llm_stats}
\end{table}

We observe from the mean estimates that the sampled data aligns most in travel time, income and party size, and displays some variation in costs, and significant variation in wait time compared to the real distribution. This can be attributed to simulator parameters, which generate much larger wait times more often than observed in real data. Similarly, the income is biased towards higher values due to the simulator not being adjusted for skewness in real income reports. 
\section{Extended baseline comparison for Power Grid Resource Allocation}\label{sec:power-ablation}

\begin{figure}
    \centering
    \includegraphics[width=0.8\linewidth]{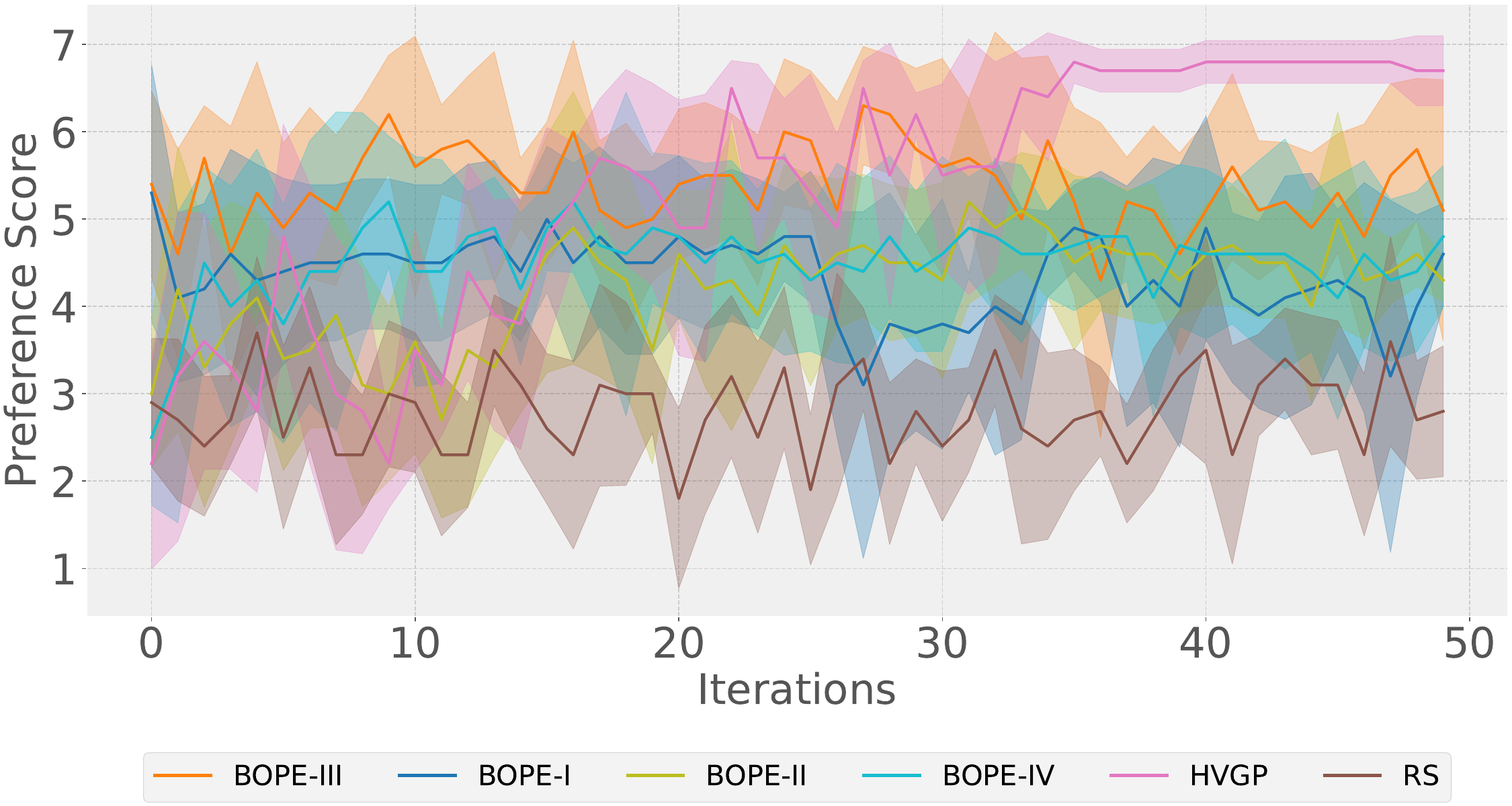}
    \caption{\textbf{Extended comparison with baselines (Case-5)}. Comparison against \baselinename{BOPE} based approaches, plotted against \baselinename{Random sampling} for reference.}
    \label{fig:power-ablation}
\end{figure}

In this section we discuss extended baseline comparison with \baselinename{BOPE} based approaches adopted from~\cite{lin2022preference}, that considers a composite problem setting, and decouples the two stages of \textit{preference exploration} and \textit{experimentation}. The original approach is open-ended, with a lot of problem-specific flexibility on how the various design stages can be chosen. We consider four ablations of their technique, and compare against our joint acquisition approach with HVGP. \Cref{fig:power-ablation} shows the comparison on Power Grid Resource Allocation case study (\Cref{subsec:power}), Case-5 problem, using preference score as a metric of evaluation. 

\begin{itemize}
    \item \baselinename{BOPE-I}: A two-phase strategy that begins with qEUBO (utility-focused exploration,\citep{astudillo2019bayesian}) and switches to qNEI in the second half (objective-driven refinement), useful for testing the effect of premature exploitation.

    \item \baselinename{BOPE-II}: A two-phase strategy that begins with qNEI (to explore the objective space) and switches to qEUBO in the second half of optimization (to exploit the learned preference model), using a frozen outcome snapshot for qEUBO.
    \item \baselinename{BOPE-III}:A qEUBO-only variant where experiment selection uses qEUBO with a new objective realization sampled at every iteration instead of a frozen snapshot, encouraging higher exploration through objective variation.
    \item \baselinename{BOPE-IV}: A baseline BOPE variant that uses standard preference exploration (EUBO-$\tau$) and selects experiments exclusively with qNEI, fully refitting both objective and subjective GPs after each update. 
\end{itemize}

\paragraph{Results.} We observe that our method explores higher preference regions after some initial exploration, unlike \baselinename{BOPE} variants, where the performance is highly sensitive to design choices. \baselinename{BOPE-I} starts off with higher preference values due to optimization driven approach, but fails to handle both optimization and exploration, leading to loss of convergence in second half. This also shows that the performance is sensitive to the point where the design stages are switched. \baselinename{BOPE-II} explores first, and then performs optimization, leading to better performance eventually, however, as in \baselinename{BOPE-I}, the complete disconnect bewteen exploration and exploitation leads to sub-optimal performance. \baselinename{BOPE-III} combines both exploration and exploitation but is still more exploitation centric, leading to higher preference scores than \baselinename{BOPE-I,II} but sub-optimal compared to \baselinename{HVGP}. \baselinename{BOPE-IV} and our method \baselinename{HVGP} have similar trends, due to similar concept of refitting both models at each stage, but \baselinename{BOPE-IV} converges to a sub-optimal value. Our acquisition strategy combines information from both objective and subjective layers at once, which leads to higher sample efficiency, and discovery of higher preference region.

\end{document}